\documentclass[letterpaper, 10 pt, conference]{ieeeconf}  % Comment this line out if you need a4paper

\IEEEoverridecommandlockouts                              % This command is only needed if 
\usepackage{graphics} % for pdf, bitmapped graphics files
\usepackage{epsfig} % for postscript graphics files
\usepackage{times} % assumes new font selection scheme installed
\usepackage{amsmath} % assumes amsmath package installed
\usepackage{amssymb}  % assumes amsmath package installed
\usepackage{booktabs}
\usepackage{caption}
\usepackage{graphicx}
\usepackage{float} 
\usepackage{hyperref}
\usepackage{subcaption}
\usepackage{color}
\usepackage{cleveref}
\usepackage{multirow}
\usepackage{xcolor} 
\usepackage[table]{xcolor}

\definecolor{first}{rgb}{0.75, 0.92, 0.75}      % 绿色 (1st)
\definecolor{second}{rgb}{0.82, 0.93, 0.60}      % 更绿的黄绿 (2nd)
\definecolor{third}{rgb}{1.00, 0.97, 0.65}      % 黄色 (3rd)
\definecolor{annobg}{rgb}{0.90,0.95,1.00}  % 淡灰，用于区块背景

\hypersetup{
    colorlinks=true,
    linkcolor=blue,
    filecolor=magenta,      
    urlcolor=cyan,
    pdftitle={Overleaf Example},
    pdfpagemode=FullScreen,
    }

\newcommand{\boldparagraph}[1]{\vspace{0.2cm}\noindent{\bf #1:} }
\DeclareMathAlphabet{\mathbbold}{U}{bbold}{m}{n}

\title{\LARGE \bf
UnIRe: Unsupervised Instance Decomposition for Dynamic Urban Scene Reconstruction
}
\author{
Yunxuan Mao, Rong Xiong, Yue Wang, Yiyi Liao*\\
Zhejiang University\\
{\tt\small maoyunxuan@zju.edu.cn,  rxiong@zju.edu.cn, ywang24@zju.edu.cn, yiyi.liao@zju.edu.cn}
}

\begin{document}
\twocolumn[{%
\renewcommand\twocolumn[1][]{#1}%
\maketitle
\thispagestyle{empty}
\pagestyle{empty}
\begin{center}
    \centering
    \captionsetup{type=figure}
    \includegraphics[width=\textwidth]{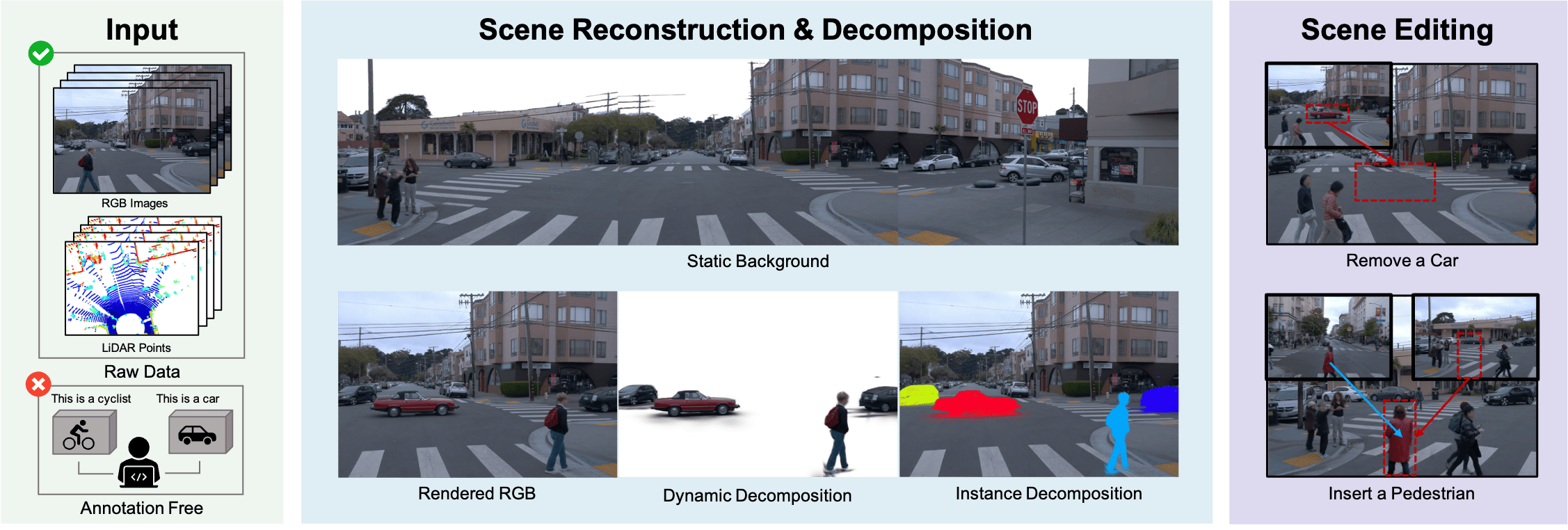}
    \captionof{figure}{\textbf{UnIRe.}  Our method enables dynamic urban scene reconstruction and decomposition without requiring manual annotation. (a) \textbf{UnIRe} separates static and dynamic components while achieving instance-aware decomposition of dynamic objects. (b) \textbf{UnIRe} also supports scene editing and simulation applications, such as removing a vehicle or adding a pedestrian.}
\end{center}%
}]
% \maketitle

% \begin{figure*}
%     \centering
%     \includegraphics[width=\linewidth]{ICCV2025-Author-Kit/img/teaser.png}
%     \caption{Caption}
%     \label{teaser}
% \end{figure*}

\begin{abstract}
Reconstructing and decomposing dynamic urban scenes is crucial for autonomous driving, urban planning, and scene editing. However, existing methods fail to perform instance-aware decomposition without manual annotations, which is crucial for instance-level scene editing.
We propose UnIRe, a 3D Gaussian Splatting (3DGS) based approach that decomposes a scene into a static background and individual dynamic instances using only RGB images and LiDAR point clouds. At its core, we introduce 4D superpoints, a novel representation that clusters multi-frame LiDAR points in 4D space, enabling unsupervised instance separation based on spatiotemporal correlations. These 4D superpoints serve as the foundation for our decomposed 4D initialization, i.e., providing spatial and temporal initialization to train a dynamic 3DGS for arbitrary dynamic classes without requiring bounding boxes or object templates.
Furthermore, we introduce a smoothness regularization strategy in both 2D and 3D space, further improving the temporal stability.
Experiments on benchmark datasets show that our method outperforms existing methods in decomposed dynamic scene reconstruction while enabling accurate and flexible instance-level editing, making it a practical solution for real-world applications. 
\end{abstract}

\section{Introduction}
\label{sec:intro}

The reconstruction and decomposition of dynamic urban scenes is  crucial for applications such as autonomous driving, urban planning, and scene simulation and editing. Recent advances in 3D Gaussian Splatting (3DGS)~\cite{kerbl20233dgs} have significantly improved scene reconstruction quality, enabling high-fidelity representations using only 2D supervision. 
However, achieving instance-level decomposition in dynamic urban scene reconstruction for arbitrary object classes, such as pedestrians and vehicles, remains a significant challenge.
%However, adapting 3DGS to dynamic urban environments and enabling scene editing present three key challenges. First, the standard 3DGS is designed for static scenes and lacks an effective mechanism to model objects undergoing significant displacement. Second, 3DGS-based methods are highly sensitive to initialization, and poor motion priors can lead to unstable optimization, especially in dynamic scenes. Third, to enable scene editing and simulation, such as removing or inserting objects, it is essential to decompose the scene into independent dynamic instances.

Recently, several approaches have been proposed to address this challenge, broadly classified into scene graph-based methods and self-supervised decomposition methods.
Scene graph-based methods~\cite{yan2024streetgs, zhou2024hugs, chen2024omnire} model dynamic scenes as structured graphs, where each instance is segmented with 3D bounding box annotations and assigned a canonical space. This formulation provides robust motion initialization and ensures instance-aware decomposition with structured 3D shapes, making it well-suited for scene editing. 
%Additionally, the scene graph deforms the canonical space with per-point motion, enabling high reconstruction quality over long sequences (\~150 frames). 
However, these methods heavily rely on manually labeled bounding boxes that are expensive to obtain, which limits their applicability across diverse urban environments.
Self-supervised decomposition methods~\cite{chen2023periodic, yang2023emernerf, turki2023suds, peng2024desire} eliminate the need for manual annotations by learning to distinguish between static backgrounds and dynamic regions directly from RGB images and LiDAR data, enhancing practicality. PVG~\cite{chen2023periodic} represents dynamic scenes using short-lived Gaussians, assigning different Gaussians to the same dynamic object at different timestamps. SplatFlow~\cite{sun2025splatflow} uses self-supervised scene flow to initialize dynamic Gaussians. %However, these methods require significantly more Gaussians to maintain reconstruction quality in long sequences, leading to increased memory consumption and degraded rendering efficiency. 
However, these methods lack a canonical space for each dynamic instance, making it hard to merge information observed across frames. In addition, these methods only decompose static and dynamic regions, without further decomposing dynamic instances, making object-level editing challenging.
%Additionally, they don't have a structured 3D shape for objects and lack per-instance decomposition, making object-level editing challenging.

%Given these limitations, we aim to develop a method that maintains high-fidelity scene reconstruction while achieving instance-aware decomposition without relying on extra annotations, enabling instance-level scene editing.
%
In this work, we propose UnIRe, a 3DGS-based framework that decomposes a scene into a static background and individual dynamic instances using only RGB images and LiDAR point clouds. 
At its core, UnIRe introduces 4D superpoints (akin to superpixels), a novel representation that clusters multi-frame LiDAR points in 4D space. We first generate over-segmented 4D superpoints by propagating per-frame clustering results using self-supervised flow estimation. Next, we cluster these 4D superpoints leveraging their spatiotemporal correlations, achieving instance-level decomposition for arbitrary dynamic classes, without the need for bounding boxes or object templates.
This decomposition serves as the initialization for the canonical space and per-point deformation of the dynamic 3DGS.
Furthermore, to prevent overfitting and unstable motion, we introduce a smoothness regularization strategy in both 2D and 3D, improving the motion consistency across different frames. Together, these components enable high-fidelity rendering and instance-aware decomposition, enabling flexible scene editing.
%
% In this work, we introduce GUIDE, a self-supervised framework for urban scene reconstruction and instance-aware decomposition using Gaussian Splatting. GUIDE requires only RGB images and 3D LiDAR point clouds as input, decomposing the scene into a static background and independent dynamic instances while ensuring spatiotemporal consistency without relying on additional annotations.
% At the core of GUIDE is 4D SuperPoint-Based Initialization, where we introduce 4D SuperPoint, a compact spatiotemporal representation that aggregates self-supervised scene flow across frames to achieve temporally consistent decomposition of static and dynamic components. This decomposition provides a structured initialization, including a canonical space and per-point deformation, for Gaussian Splatting-based scene reconstruction, where static and dynamic components are modeled separately. To further enforce temporal consistency, we introduce a smoothness regularization strategy that refines motion coherence across frames, ensuring stable and editable reconstructions. These components collectively enable GUIDE to achieve high-fidelity rendering and instance-aware decomposition.
%
Experiments on Waymo~\cite{sun2020waymo} and KITTI~\cite{Geiger2012kitti} datasets demonstrate that UnIRe achieves state-of-the-art performance in an annotation-free manner while enabling instance-level editing.

\begin{figure*}[ht]
    \centering
    \includegraphics[width=\linewidth]{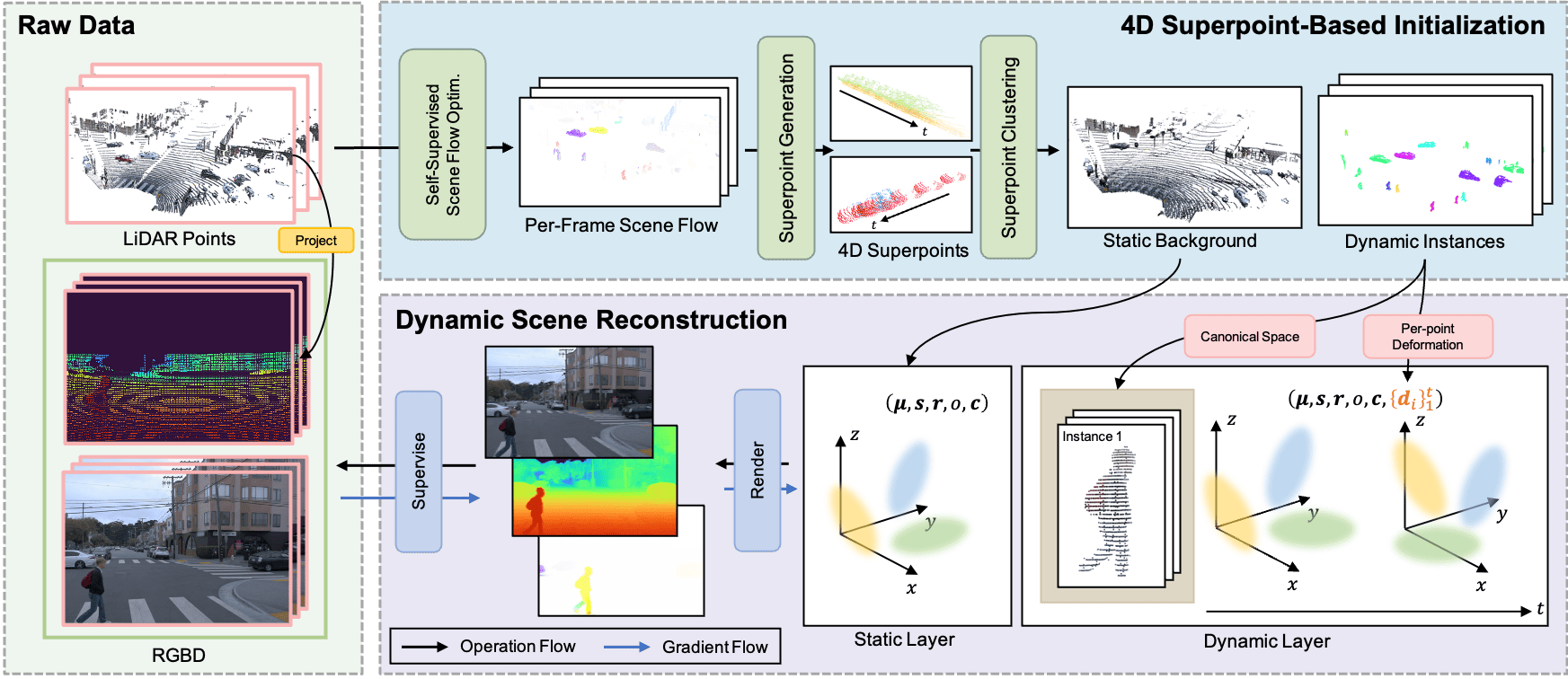}
    \vspace{-0.5cm}
    \caption{\textbf{Method Overview.}  Our method consists of two core components: 4D SuperPoint-Based Initialization and Dynamic Scene Representation. 
            4D SuperPoint-Based Initialization takes LiDAR points as input and estimates scene flow using a self-supervised optimization method. 
            Then, 4D SuperPoint Generation and 4D SuperPoint Clustering decompose the scene into a static background and dynamic instances. 
            The dynamic instances are further used to construct a canonical space and per-point deformation $\{d_i\}_1^t$. 
            Dynamic Scene Representation utilizes the static background to initialize the static layer, while the canonical space and per-point deformation serve as the initialization for the dynamic layer. 
            The model is supervised by ground truth images and depth maps projected from LiDAR points.}
    \label{overview}
\end{figure*}

\section{Related Work}
\label{sec:relatedwork}

\boldparagraph{Dynamic Scene Reconstruction}
Neural scene representations~\cite{mildenhall2021nerf, barron2021mip, barron2022mip360, muller2022instant, kerbl20233dgs, huang20242dgs} have significantly advanced novel view synthesis, inspiring extensive research in dynamic scene reconstruction. NeRF-based methods~\cite{park2021nerfies, park2021hypernerf, pumarola2021dnerf} rely on neural deformation fields and canonical spaces to model motion but lack explicit geometry, limiting their applicability to large-scale, real-world urban environments. Similarly, recent works on 3D Gaussian Splatting (3DGS)~\cite{yang2024deformable, wu20244d} employ neural deformation-based motion modeling. However, neural deformation fields are inadequate for capturing large-scale dynamic variations.

To overcome these limitations, recent approaches leverage the explicit nature of 3DGS to represent per-point motion. One strategy extends the spatial distribution of Gaussian points into a four-dimensional space~\cite{yang20234dgs, duan20244drotor}, embedding temporal variations directly into Gaussian parameters. This formulation models the same object with different Gaussians at different time steps, leading to increased memory consumption in large scenes and inconsistent motion. Another strategy employs per-point deformation, where each Gaussian is associated with a canonical space to maintain temporal consistency~\cite{luiten2024dynamic, wang2024shape, lin2024Gaussian}.

\boldparagraph{Urban Scene Reconstruction and Decomposition}
Urban scene reconstruction methods can be broadly categorized into annotation-dependent scene graph methods and self-supervised scene decomposition methods. Annotation-dependent methods construct a scene graph where each object is explicitly decomposed into instances using 3D bounding box annotations, enabling dynamic scene representation with instance decomposition. Methods such as MARS~\cite{wu2023mars}, UniSim~\cite{yang2023unisim}, DrivingGaussian~\cite{zhou2024drivingGaussian}, StreetGS~\cite{yan2024streetgs}, OmniRe~\cite{chen2024omnire}, and HUGS~\cite{zhou2024hugs} follow this paradigm. The scene graph served as an instance-aware canonical space, making object-level editing easy. However, they rely on manually labeled 3D bounding boxes~\cite{wu2023mars,yang2023unisim,chen2024omnire} or accurate 3D tracking initialization~\cite{zhou2024drivingGaussian}, limiting their scalability across diverse urban environments.

In contrast, self-supervised decomposition methods eliminate the need for annotations by learning to separate static and dynamic components during training. EmerNeRF~\cite{yang2023emernerf} and SUDS~\cite{turki2023suds} estimate motion using implicit flow fields, constraining scene dynamics via multi-frame optimization. While these NeRF-based methods improve scalability, they often struggle with slow rendering speeds and limited reconstruction quality. PVG~\cite{chen2023periodic} and DeSiReGS~\cite{peng2024desire} directly embed temporal variations into Gaussian representations, enabling motion-aware reconstruction without explicit deformation fields. 
%However, these methods require significantly more Gaussians to maintain reconstruction quality in long sequences, leading to increased memory consumption and degraded rendering efficiency. 
Recently,SplatFlow \cite{sun2025splatflow} leverage scene flow to initialize Gaussians in dynamic scenes. However, these methods lack explicit per-instance decomposition and canonical space, making scene editing challenging.

\section{Preliminaries}

\boldparagraph{3D Gaussian Splatting} 
3D Gaussian Splatting~\cite{kerbl20233dgs} (3DGS) represents a scene as a collection of learnable anisotropic Gaussians, $\mathcal{G} = \{ g \}$. Each Gaussian $g = (\boldsymbol{\mu}, \boldsymbol{s}, \boldsymbol{r}, o, \boldsymbol{c} )$ is parameterized by the following attributes: a position center $\boldsymbol{\mu} \in \mathbb{R}^3$, a scaling vector $\boldsymbol{s}$, a quaternion $\boldsymbol{r} \in \mathbb{R}^{4}$, an opacity scalar $o$, and a color vector $\boldsymbol{c}$, which is represented using spherical harmonics. The spatial distribution of each 3D Gaussian is given by:
\begin{equation}
    G(\boldsymbol{x}) = \exp \left\{ -\frac{1}{2} (\boldsymbol{x} - \boldsymbol{\mu})^\top \Sigma^{-1} (\boldsymbol{x} - \boldsymbol{\mu}) \right\}.
\end{equation}
The covariance matrix is $
    \Sigma = \mathbf{R} \mathbf{S} \mathbf{S}^\top \mathbf{R}^\top$
, where $\mathbf{S}$ is a diagonal scaling matrix and $\mathbf{R}$ is a rotation matrix, parameterized by the scaling vector $\boldsymbol{s}$ and the quaternion $\boldsymbol{r}$. 

To render an image from a given viewpoint, the 3D Gaussian ellipsoids are projected onto a 2D image plane, forming 2D ellipses. The projected Gaussians are sorted in depth order, and the pixel color is obtained via alpha blending:
\begin{equation}
    C = \sum_{i \in \mathcal{N}} c_i \alpha_i \prod_{j=1}^{i-1} (1 - \alpha_j),
\end{equation}
where $\alpha_i$ and $c_i$ denote the opacity and color of the $i$-th Gaussian derived from the learned opacity and spherical harmonics (SH) coefficients of the corresponding Gaussian.

% \noindent 
% \textbf{Self-Supervised Scene Flow Optimization.}  
% Scene flow estimation is the task of predicting per-point 3D motion between consecutive LiDAR frames. Given a point cloud at time $t$, denoted as $\mathcal{P}^t = \{ \mathbf{p}_i^t \in \mathbb{R}^3 \}$, the scene flow function $\mathbf{F}: \mathbb{R}^3 \to \mathbb{R}^3$ estimates the displacement of each point $\mathbf{p}_i^t$ to its next position $\mathbf{p}_i^{t+1}$:

% \begin{equation}
%     \mathbf{p}_i^{t+1} = \mathbf{p}_i^t + \mathbf{F}(\mathbf{p}_i^t).
% \end{equation}

% Scene flow is inherently pairwise and only models short-term correspondences between two consecutive frames. Traditional approaches rely on ground-truth annotations, which are costly to obtain. Self-supervised methods optimize flow by enforcing geometric consistency. Specifically, \textit{Let It Flow}~\cite{vacek2024let} provides an effective self-supervised approach for estimating scene flow without requiring manual annotations.

% While these methods estimate scene flow at a pairwise level, our approach extends this by leveraging the entire sequence of flow predictions to construct a 4D initialization of dynamic urban scenes in \textbf{GUIDE}.

% You must include your signed IEEE copyright release form when you submit your finished paper.
% We MUST have this form before your paper can be published in the proceedings.

% Please direct any questions to the production editor in charge of these proceedings at the IEEE Computer Society Press:
% \url{https://www.computer.org/about/contact}.
\section{Method}

%Our method is designed for dynamic urban scene reconstruction and editing using only RGB images and LiDAR data without requiring any extra annotations. A key requirement for scene editing is the ability to independently manipulate dynamic objects, which requires decomposing dynamic instances and establishing a canonical representation as a consistent reference frame across time. As discussed in our supplementary add reference to the supplementary~\cref{sec:supp_discuss}, existing 2D RGBs or 3D LiDAR based detection and tracking methods struggle to provide robust results in a label-free manner. Therefore, we propose a simple yet effective unsupervised method for decomposing the scene based on the spatial-temporal correlation of LiDAR points.
Our method enables dynamic urban scene reconstruction and editing using only RGB images and LiDAR data, without additional annotations. A key requirement for scene editing is the ability to independently manipulate dynamic objects, which requires decomposing dynamic instances and aggregating cross-time information in canonical space. Existing 2D RGB and 3D LiDAR-based detection and tracking methods struggle to deliver robust results in a label-free manner. To address this, we propose a simple yet effective unsupervised approach for scene decomposition based on the spatial-temporal correlation of LiDAR points.

% As shown in \cref{overview}, our method consists of two core components: 4D superpoint based initialization and Dynamic Scene Representation. 4D superpoint based initialization encodes both spatial structure and temporal motion, forming a canonical space with per-point deformation to ensure temporally consistent object identities. Building on this, Dynamic Scene Representation employs Gaussian Splatting to reconstruct the scene by decomposing it into static and dynamic components, enabling efficient motion modeling. Additionally, an optimization strategy enforces temporal consistency, ensuring smooth and stable motion trajectories across the sequence.

As shown in \cref{overview}, our method consists of two core components: 4D superpoint-based initialization (\cref{sec:initialization}) and dynamic scene reconstruction (\cref{sec:scene}), with the former provides spatial and temporal initialization for the latter. Our reconstruction is supervised by ground truth images and depth maps projected from LiDAR points, combined with smooth regularization to improve temporal consistency (\cref{sec:optimization}).
%The 4D superpoint-based initialization takes LiDAR points as input and and first estimates scene flow using a self-supervised optimization method. Then, 4D superpoint Generation and 4D superpoint Clustering decompose the scene into a static background and dynamic instances. The dynamic instances are further used to construct a canonical space and per-point deformation. Dynamic Scene Representation utilizes the static background to initialize the static layer, while the canonical space and per-point deformation serve as the initialization for the dynamic layer. The model is supervised by ground truth images and depth maps projected from LiDAR points.

\subsection{4D Superpoint-Based Initialization}
\label{sec:initialization}

\begin{figure}
    \centering
    \includegraphics[width=\linewidth]{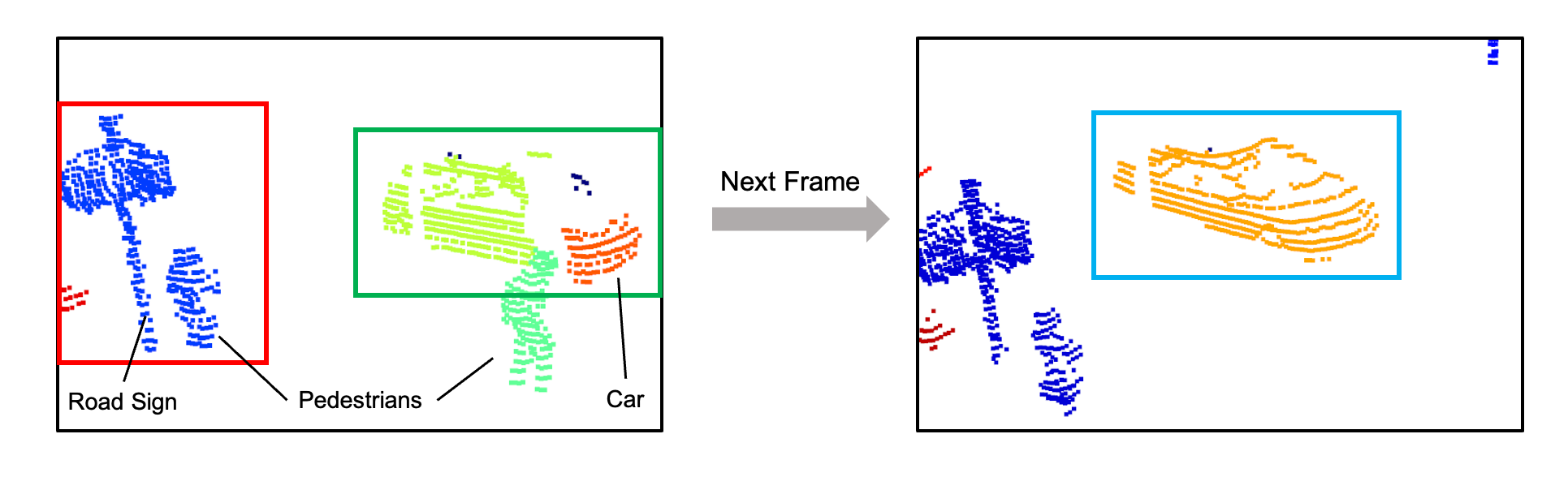}
    \vspace{-0.5cm}
    \caption{\textbf{Visualization of Decomposition Challenges.} {Under-Decomposition} ({\color{red}red box}), {Over-Decomposition} ({\color{teal}green box}), and {Inconsistent Cluster IDs} ({\color{cyan}cyan box}).}
    \label{seg}
\end{figure}

%To establish a canonical space, we need a temporally consistent decomposition of dynamic instances across the sequence.
We propose a simple method that decomposes a sequence of LiDAR points based on unsupervised clustering.
As shown in \cref{seg}, per-frame clustering-based decomposition can be inconsistent due to object motion and occlusions, leading to three major challenges:

\begin{itemize}
    \item {Inconsistent Cluster IDs}: The same object may be assigned different cluster IDs across frames.
    \item {Over-Decomposition}: A single object may be divided into multiple clusters due to temporary occlusions or variations in the density of LiDAR points.
    \item {Under-Decomposition}: Spatially close but distinct objects may be mistakenly merged into the same cluster.
\end{itemize}

\begin{figure*}
    \centering
    \begin{subfigure}{0.24\linewidth}
    \centering
        \includegraphics[width=\linewidth]{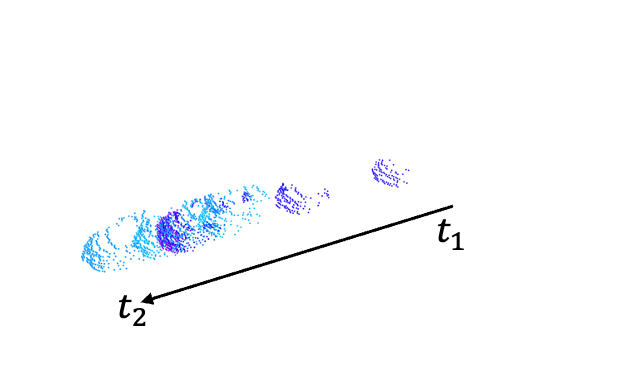}
        \caption{DBSCAN}
        \label{subfig:dbscan}
    \end{subfigure}
    \begin{subfigure}{0.24\linewidth}
    \centering
        \includegraphics[width=\linewidth]{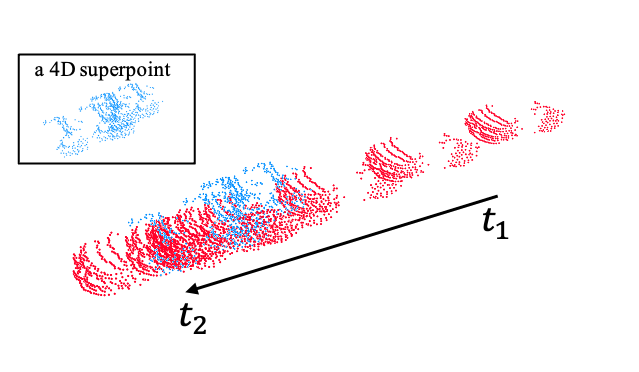}
        \caption{4D Superpoint Generation}
        \label{subfig:4dgene}
    \end{subfigure}
    \begin{subfigure}{0.24\linewidth}
    \centering
        \includegraphics[width=\linewidth]{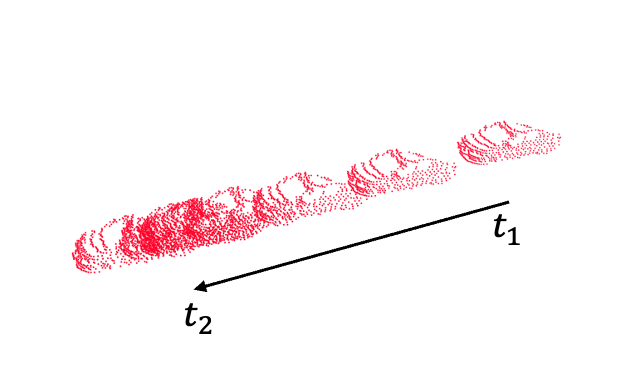}
        \caption{4D Superpoint Clustering}
        \label{subfig:4dclus}
    \end{subfigure}
    \begin{subfigure}{0.24\linewidth}
    \centering
        \includegraphics[width=\linewidth]{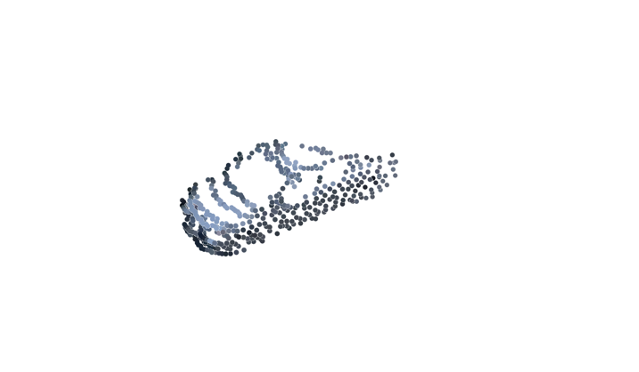}
        \caption{Canonical Space}
        \label{subfig:canonical}
    \end{subfigure}
    \vspace{-0.2cm}
    \caption{\textbf{Visualization of 4D Superpoint-Based Initialization.} To provide an intuitive understanding, we take a car as an example. First, DBSCAN is applied independently to each frame, resulting in inconsistent clustering. Then, we align clusters across frames and form temporally consistent 4D superpoints (different 4D superpoints are in different colors). Finally, these 4D superpoints are clustered to achieve instance-level decomposition and establish a canonical space.}
    \label{fig:4dsuperpoint}
\end{figure*}

To resolve these challenges, we introduce \textbf{4D superpoint}, a spatiotemporal representation that ensures consistent instance decomposition throughout the sequence. Formally, a 4D superpoint is defined as a cluster of points over a sequence of time:
\begin{equation}
\mathcal{S} = \{\mathcal{C}^t\}_{t=1}^{T},
\end{equation}
where $\mathcal{C}^t$ is the cluster of points at frame $t$. The pipeline of 4D superpoint based initialization is shown in~\cref{fig:4dsuperpoint}.

\boldparagraph{4D Superpoint Generation} 
Given a sequence of LiDAR points $\{\mathcal{P}^t=\{\mathbf{p}_i^t\}\}_1^T$, we pre-process each frame by removing ground points to stabilize clustering results. Then, we apply DBSCAN~\cite{ester1996density} independently to each frame to group points into clusters. Next, we apply a self-supervised scene flow optimization method \textit{let it flow}~\cite{vacek2024let} to estimate the scene flow between every two adjacent frames, denoted as $\{\mathcal{SF}^t=\{\mathbf{f}_i^t\}\}_1^{T-1}$.

%Instead of directly propagating cluster IDs point by point, 
Next, we establish correspondences between DBSCAN-generated clusters across frames. Given a set of clusters $\{\mathcal{C}^t_k\}_{k=1}^{K_t}$ and $\{\mathcal{C}^{t+1}_m\}_{m=1}^{K_{t+1}}$ in frame $t$ and $t+1$, we match clusters based on scene flow consistency. Specifically, for each cluster $\mathcal{C}^t_k$, we find the most likely corresponding cluster in frame $t+1$ by maximizing the number of points that remain spatially aligned after applying scene flow:
\begin{equation}
\mathcal{C}^{t+1}_{\text{match}} = \arg\max_{\mathcal{C}^{t+1}_m} \sum_{\mathbf{p}_i^t \in \mathcal{C}^t_k} \mathbbold{1} (\mathbf{p}_i^t + \mathbf{f}_i^t),
\end{equation}
where $\mathbbold{1}(\cdot)$ is an indicator function that counts the number of points in $\mathcal{C}^t_k$ whose scene flow displacement lands within $\mathcal{C}^{t+1}_m$. The cluster ID of $\mathcal{C}^t_k$ is then assigned to $\mathcal{C}^{t+1}_{\text{match}}$, ensuring that cluster IDs remain consistent across frames.

However, due to occlusions, motion variations, and object interactions, clusters may undergo three types of transformations: vanishing, emergence, and splitting:
\begin{itemize}
\item Vanishing: If $\mathcal{C}^t_k$ has no valid match in $t+1$, it is registered as vanishing and removed from further tracking.
\item Emergence: If a cluster $\mathcal{C}^{t+1}_m$ has no corresponding cluster in $t$, it is registered as a new cluster.
\item Splitting: If a cluster $\mathcal{C}^t_k$ is matched with multiple clusters in $t+1$, it is divided into multiple clusters, and each new cluster maintains its own separate identity.
\end{itemize}

After applying these alignment rules across the sequence, we obtain a set of 4D superpoints, each 4D superpoint corresponds to a cluster that maintains a consistent identity over time: $\mathcal{S}_k = \{\mathcal{C}_k^t\}_{t=t_1}^{t_2}$, where $\mathcal{C}_k^t$ denotes the cluster state at time $t$, spanning frames $t_1$ to $t_2$.

However, our cluster splitting can lead to over-decomposition, where a single object is unnecessarily divided into multiple clusters, as shown in~\cref{subfig:4dgene}. To mitigate this issue, we cluster 4D superpoint in the next step by leveraging spatiotemporal similarity.

\boldparagraph{4D Superpoint Clustering}  
Over-decomposition from the splitting process results in a single object being divided into multiple 4D superpoints due to inconsistencies in motion and occlusion. To refine these results, we leverage spatiotemporal similarity to cluster 4D superpoints into consistent instances while preserving distinct object identities.

We first estimate the spatiotemporal properties of each 4D superpoint, including its position and motion in each frame. Given a 4D superpoint $\mathcal{S}_k$ at time $t$, we compute:
\begin{equation} \boldsymbol{\mu}_k^t = \frac{1}{|\mathcal{S}_k^t|} \sum_{\mathbf{p}_i^t \in \mathcal{S}_k^t} \mathbf{p}_i^t, \quad
\mathbf{F}_k^t = \frac{1}{|\mathcal{S}_k^t|} \sum_{\mathbf{p}_i^t \in \mathcal{S}_k^t} \mathbf{f}_i^t, 
\label{eq:flow}
\end{equation}
where $\boldsymbol{\mu}_k^t$ represents the spatial centroid of the 4D superpoint, while $\mathbf{F}_k^t$ denotes its average scene flow. 

Then, we compute the spatiotemporal similarity matrix $\mathcal{M}$ of the 4D superpoints, which integrates both motion direction and spatial proximity. Given two 4D superpoints $\mathcal{S}_k$ and $\mathcal{S}_l$ in frame $t$, we define their similarity as
\begin{equation}
     \mathcal{M}^t_{k,l} = \lambda \frac{\mathbf{F}_{k}^t \cdot \mathbf{F}_{l}^t}{|\mathbf{F}_{k}^t| |\mathbf{F}_{l}^t|} + (1-\lambda) \exp \left( -\frac{\|\boldsymbol{\mu}_{k}^t - \boldsymbol{\mu}_{l}^t\|^2}{\sigma^2} \right),
\end{equation}
where the first term measures the motion direction similarity using cosine similarity, while the second term captures the spatial proximity, and $\lambda$ controls the balance between motion and spatial similarity. The final spatiotemporal similarity matrix is obtained by aggregating frame-wise similarities $\mathcal{M}=\sum_t\mathcal{M}^t$. 

Finally, we apply DBSCAN to $\mathcal{M}$, which merges overdecomposed 4D superpoints, producing a temporally consistent decomposition for dynamic objects, as shown in~\cref{subfig:4dclus}. The clustering result is used as the instance decomposition, denoted as $\mathcal{I}=\{\mathcal{I}^t\}_{t=1}^T$.

\boldparagraph{Canonical Space Initialization}  
For each dynamic instance, we establish a shared canonical shape for initializing the dynamic 3DGS.
%To establish a stable canonical space for dynamic instances, 
Specifically, we select a reference frame that provides the most complete observation of each instance. Given an instance $\mathcal{I}_k$ tracked from $t_1$ to $t_2$, we define its reference frame $t^*$ as the frame that contains the maximum number of points:
\begin{equation}
    t^* = \arg \max_{t \in [t_1, t_2]} |\mathcal{I}^t_k|,
\end{equation}
where $\mathcal{I}^t_k$ denotes the set of points belonging to instance $\mathcal{I}_k$  at time $t$. This ensures that the canonical space is defined based on the most complete observation of the instance. 
% The complete canonical space is then represented as:
% \begin{equation}
%     \mathbb{C} = \bigcup_k \mathcal{I}^{t^*}_k.
% \end{equation}

\boldparagraph{Per-point Deformation Initialization}  
After obtaining the canonical space, we compute the per-point deformation for each point in the scene, serving as temporal initialization for the dynamic 3DGS. The per-point deformation $\mathbf{d}_i^t$ at time $t$ is defined as:
\begin{equation}
    \mathbf{d}_i^t = (\delta^t_\mathbf{x_i}, \delta^t_\mathbf{s_i}, \delta^t_\mathbf{r_i}) = (\sum_{\tau=t^*}^{t} \mathbf{F}^\tau(\mathcal{I}(\mathbf{p}_i)), 0, 0),
\end{equation}
where $\delta^t_\mathbf{x_i}, \delta^t_\mathbf{s_i}, \delta^t_\mathbf{r_i}$ are of set $\mathbf{F}^\tau(\mathcal{I})$ represents the cumulative estimated scene flow of instance $\mathcal{I}$ up to time $\tau$, and $\mathcal{I}(\mathbf{p}_i)$ denotes the instance ID of point $\mathbf{p}_i$.

\subsection{Dynamic Scene Reconstruction}
\label{sec:scene}

After initializing the 3D canonical space and per-point deformations, we build our dynamic scene reconstruction framework by integrating these motion priors into 3D Gaussian Splatting (3DGS). The scene is decomposed into a static layer and a dynamic layer, where instances are classified by thresholding the average scene flow magnitude. Static instances form a set of Gaussians $\mathcal{G}^{static}$, while dynamic instances are represented in a canonical space with per-point deformations applied to each Gaussian.

The complete representation is $\mathcal{G}=(\mathcal{G}^{static}, \mathcal{G}^{dynamic})$, where learnable parameters include Gaussian attributes and deformation fields. To enforce temporal consistency, we  render optical flow between frames, using the projected Gaussian centers at two timestamps to compute $\mathbf{f}_i$, and composing them into a dense flow map $\mathcal{F}$ for smoothness regularization.

\subsection{Loss Functions}
% \subsubsection{Temporal Smooth Regularization}
\label{sec:optimization}

%To optimize both scene reconstruction and motion estimation, we perform optimization with following objective function:
Our dynamic 3DGS model is trained using RGB images and depth maps projected from LiDAR. However, we found that per-point deformation tends to overfit the training views, causing unstable motion during novel view synthesis. To improve generalization, we introduce two smoothness regularization terms in 2D and 3D space. These terms ensure temporal consistency and spatial coherence in dynamic object motion, promoting stable and smooth trajectories across different viewpoints.

\boldparagraph{2D Smoothness Regularization}
2D smoothness regularization is commonly used in unsupervised optical flow methods~\cite{ren2017unsupervised, meister2018unflow,wang2018occlusion, jonschkowski2020matters, jonschkowski2020matters}, where it helps enforce spatial coherence by penalizing abrupt motion changes. Inspired by these methods, we introduce a similar loss in UnIRe to mitigate overfitting to training views and improve motion consistency in novel view synthesis. Specifically, we define the first-order smoothness loss following~\cite{jonschkowski2020matters} as:
\begin{equation}
    \begin{split}
        \mathcal{L}_{\text{smooth}}^{\text{2D}} = & \frac{1}{N} \sum_{i=1}^{N}  
         \exp \left( -\lambda \sum_{c} \left| \frac{\partial I_c}{\partial x} \right| \right) 
        \left| \frac{\partial \mathcal{F}}{\partial x} \right| \\
        & + \exp \left( -\lambda \sum_{c} \left| \frac{\partial I_c}{\partial y} \right| \right) 
        \left| \frac{\partial \mathcal{F}}{\partial y} \right|,
    \end{split}
\end{equation}
where $I_c$ denotes the image intensity for color channel $c$, and $\lambda$ controls the edge-aware weighting.

\boldparagraph{3D Smoothness Regularization} 
While 2D smoothness regularizes optical flow in image space, it does not guarantee consistency in the 3D deformation field. To enhance spatial coherence and suppress abrupt local motion, we introduce a 3D smoothness term that enforces local velocity consistency of per-point deformation.

Specifically, the velocity $\mathbf{v}i$ of each Gaussian $\mathcal{G}i$ is constrained to be close to the mean velocity of its $K$ nearest neighbors:
\begin{equation}
\mathcal{L}{\text{smooth}}^{3D} = \frac{1}{N}\sum{i=1}^{N}\Big|\mathbf{v}_i - \tfrac{1}{K}\sum_k \mathbf{v}_j\Big|^2,
\end{equation}
where $\mathbf{v}_i = \mathbf{d}_i^{t+1} - \mathbf{d}_i^t$.

By combining 2D flow smoothness with 3D deformation smoothness, we mitigate overfitting to training views and improve motion stability in novel view synthesis, leading to more coherent reconstructions.

% \subsubsection{Loss Functions}

\boldparagraph{Full Training Loss}
Our full training loss is shown below:
\begin{equation}
\begin{split}
    \mathcal{L} = & \lambda_{\text{rgb}} \mathcal{L}_{\text{rgb}} + \lambda_{\text{depth}} \mathcal{L}_{\text{depth}} + \lambda_{\text{opacity}} \mathcal{L}_{\text{opacity}} \\
    & + \lambda_{\text{2s}} \mathcal{L}_{\text{smooth}}^{\text{2D}} + \lambda_{\text{3s}} \mathcal{L}_{\text{smooth}}^{\text{3D}}  + \lambda_{\text{reg}} \mathcal{L}_{\text{reg}},
\end{split}
\label{eq:loss}
\end{equation}
where $\mathcal{L}_{\text{rgb}}$ supervises rendered images using L1 and SSIM losses, $\mathcal{L}_{\text{depth}}$ aligns the scene with sparse LiDAR depth, and $\mathcal{L}_{\text{opacity}}$ regularizes the opacity of Gaussians to align with the sky model, ensuring proper separation between foreground objects and the background sky. $\mathcal{L}_{\text{reg}}$ represents various regularization terms.

\begin{table*}[ht!]
\centering
\scriptsize
\setlength{\tabcolsep}{4.5pt}
\begin{tabular}{lccccccccccccccccc}
\toprule
 & \multicolumn{7}{c}{Image Reconstruction}&&&\multicolumn{7}{c}{Novel View Synthesis} & \\ 
 \cmidrule{2-9} \cmidrule{11-18}
 & \multicolumn{3}{c}{Full Image} & \multicolumn{2}{c}{Human} & \multicolumn{2}{c}{Vehicle} &&&\multicolumn{3}{c}{Full Image} & \multicolumn{2}{c}{Human} & \multicolumn{2}{c}{Vehicle} & \\
Methods& \makebox[0.03\textwidth]{PSNR $\uparrow$} &  \makebox[0.03\textwidth]{SSIM $\uparrow$} &\makebox[0.03\textwidth]{LPIPS $\downarrow$} &\makebox[0.03\textwidth]{PSNR $\uparrow$} &  \makebox[0.03\textwidth]{SSIM $\uparrow$} &\makebox[0.03\textwidth]{PSNR $\uparrow$} &  \makebox[0.03\textwidth]{SSIM $\uparrow$} & \makebox[0.03\textwidth]{DL1 $\downarrow$}&&\makebox[0.03\textwidth]{PSNR $\uparrow$} &\makebox[0.03\textwidth]{SSIM $\uparrow$}& \makebox[0.03\textwidth]{LPIPS $\downarrow$}&\makebox[0.03\textwidth]{PSNR $\uparrow$} &  \makebox[0.03\textwidth]{SSIM $\uparrow$}&\makebox[0.03\textwidth]{PSNR $\uparrow$} &  \makebox[0.03\textwidth]{SSIM $\uparrow$} & \makebox[0.03\textwidth]{DL1 $\downarrow$} \\ 
 \midrule
 \rowcolor{annobg}
\multicolumn{18}{l}{\emph{\textbf{Annotation-based methods (require GT bounding boxes)}}} \\

HUGS~\cite{zhou2024hugs}
& 28.26 & 0.923 & 0.092 & 16.23 & 0.404 & 24.31 & 0.794 & 1.90
&& 27.65 & 0.914 & \cellcolor{third}0.097 & 15.99 & 0.378 & 23.27 & 0.748 & 2.13 \\

StreetGS~\cite{yan2024streetgs}
& 28.82 & 0.932 & \cellcolor{third}0.087
& 16.56 & 0.411
& \cellcolor{third}26.65 & \cellcolor{third}0.853 & 2.10
&& 27.19 & 0.889 & 0.099 & 16.28 & 0.376
& \cellcolor{third}23.89 & \cellcolor{second}0.775 & 2.17 \\

OmniRe~\cite{chen2024omnire}
& \cellcolor{second}34.81 & \cellcolor{second}0.956 & \cellcolor{second}0.054
& \cellcolor{second}27.56 & \cellcolor{second}0.828
& \cellcolor{second}28.91 & \cellcolor{second}0.897
& \cellcolor{first}\textbf{1.49}
&& \cellcolor{first}\textbf{33.03} & \cellcolor{first}\textbf{0.944} & \cellcolor{first}\textbf{0.060}
& \cellcolor{first}\textbf{24.20} & \cellcolor{first}\textbf{0.718}
& \cellcolor{first}\textbf{27.78} & \cellcolor{first}\textbf{0.867} & \cellcolor{first}\textbf{1.50} \\
\bottomrule
\midrule
% \midrule
 \rowcolor{annobg}
\multicolumn{18}{l}{\emph{\textbf{Annotation-free methods}}} \\
EmerNeRF~\cite{yang2023emernerf}
& 31.51 & 0.891 & 0.112 & 22.73 & 0.563 & 24.76 & 0.735 & 1.89
&& 29.53 & 0.878 & 0.139 & 21.37 & 0.483 & 21.98 & 0.619 & 1.97 \\

PVG~\cite{chen2023periodic}
& 32.61 & 0.936 & 0.103 & 24.72 & 0.712 & 24.29 & 0.760 & 1.86
&& 28.94 & 0.881 & 0.127 & 21.92 & 0.567 & 21.59 & 0.626 & 1.90 \\

DeSiRe\mbox{-}GS~\cite{peng2024desire}
& \cellcolor{third}32.71 & \cellcolor{third}0.949 & 0.103
& \cellcolor{third}24.87 & \cellcolor{third}0.731
& 24.51 & 0.787 & \cellcolor{second}{1.57}
&& \cellcolor{third}30.67 & \cellcolor{third}0.933 & 0.118
& \cellcolor{third}22.53 & \cellcolor{third}0.590
& 22.70 & 0.658 & \cellcolor{second}{1.55} \\

\textbf{Ours}
& \cellcolor{first}\textbf{35.58} & \cellcolor{first}\textbf{0.967} & \cellcolor{first}\textbf{0.053}
& \cellcolor{first}\textbf{30.44} & \cellcolor{first}\textbf{0.892}
& \cellcolor{first}\textbf{30.62} & \cellcolor{first}\textbf{0.922}
& \cellcolor{third}1.63
&& \cellcolor{second}31.56 & \cellcolor{second}0.935 & \cellcolor{second}0.074
& \cellcolor{second}22.75 & \cellcolor{second}0.640
& \cellcolor{second}24.82 & \cellcolor{third}0.769 & \cellcolor{third}1.64 \\
% OmniRe~\cite{chen2024omnire} &  34.81  &  0.956 & 0.054 &  27.56  & 0.828 &   28.91  & 0.897  &1.49&& 33.03  & \textbf{0.944} & \textbf{0.060} & 24.20 &  \textbf{0.718} &  27.78  & 0.867 & 1.50 \\
%\textbf{Ours w/ GT BBox} &$\checkmark$& \textbf{36.09} &  \textbf{0.963}  & \textbf{0.053} & \textbf{33.23} & \textbf{0.927}  & \textbf{29.59} &  \textbf{0.906} &\textbf{1.48}&& \textbf{33.05} & 0.943 &  0.065&\textbf{24.32} &0.712 &  \textbf{27.92} & \textbf{0.869} & \textbf{1.47}   \\
\bottomrule
% &$\checkmark$
\end{tabular}
\caption{\textbf{Quantitative comparison on Waymo Open Dataset.} Best results are highlighted as \colorbox{first}{\textbf{first}}, \colorbox{second}{second}, and \colorbox{third}{third}.
UniRe performs the best among annotation-free methods, and is also on par with annotation-based methods. DL1 refers to depth L1 (m).}
\label{waymo_quant}
\end{table*}

\begin{figure*}
  \centering
  % \begin{subfigure}{0.49\linewidth}
  % \centering
  %   \includegraphics[width=0.49\linewidth]{img/exp1/23/pvg.png}
  %   \includegraphics[width=0.49\linewidth]{img/exp1/327/pvg.png}
  %   \caption{PVG~\cite{chen2023periodic}}
  %   \label{fig:pvg-a}
  % \end{subfigure}
  % \begin{subfigure}{0.49\linewidth}
  % \centering
  %   \includegraphics[width=0.49\linewidth]{img/exp1/23/streetgs.png}
  %   \includegraphics[width=0.49\linewidth]{img/exp1/327/streetgs.png}
  %   \caption{StreetGS~\cite{yan2024streetgs}}
  %   \label{fig:short-streetgs}
  % \end{subfigure}

  \begin{subfigure}{0.49\linewidth}
  \centering
    \includegraphics[width=0.49\linewidth]{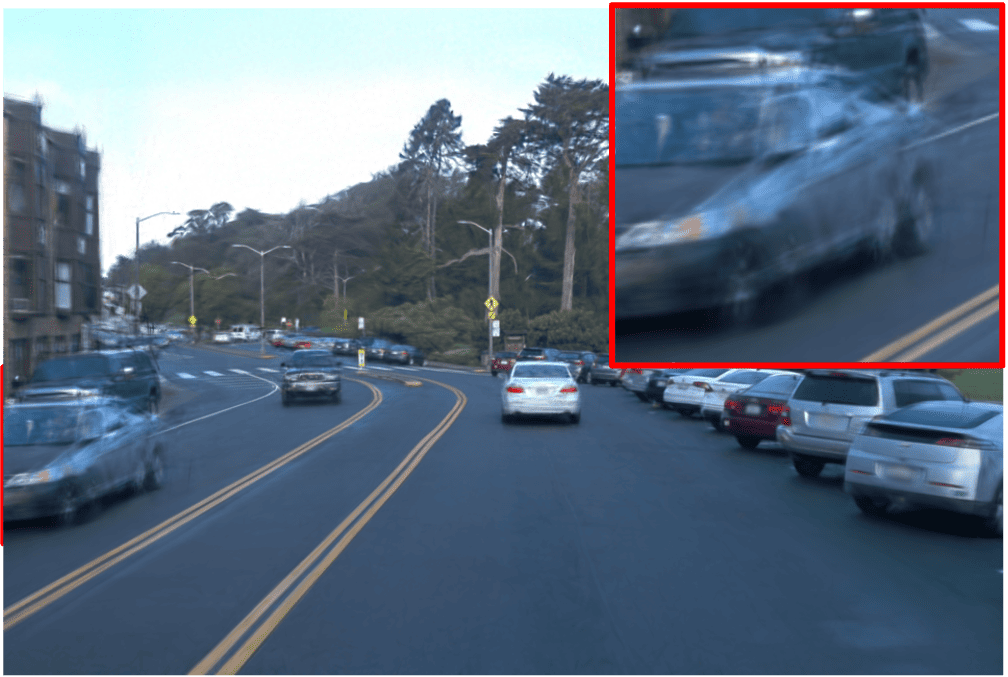}
    \includegraphics[width=0.49\linewidth]{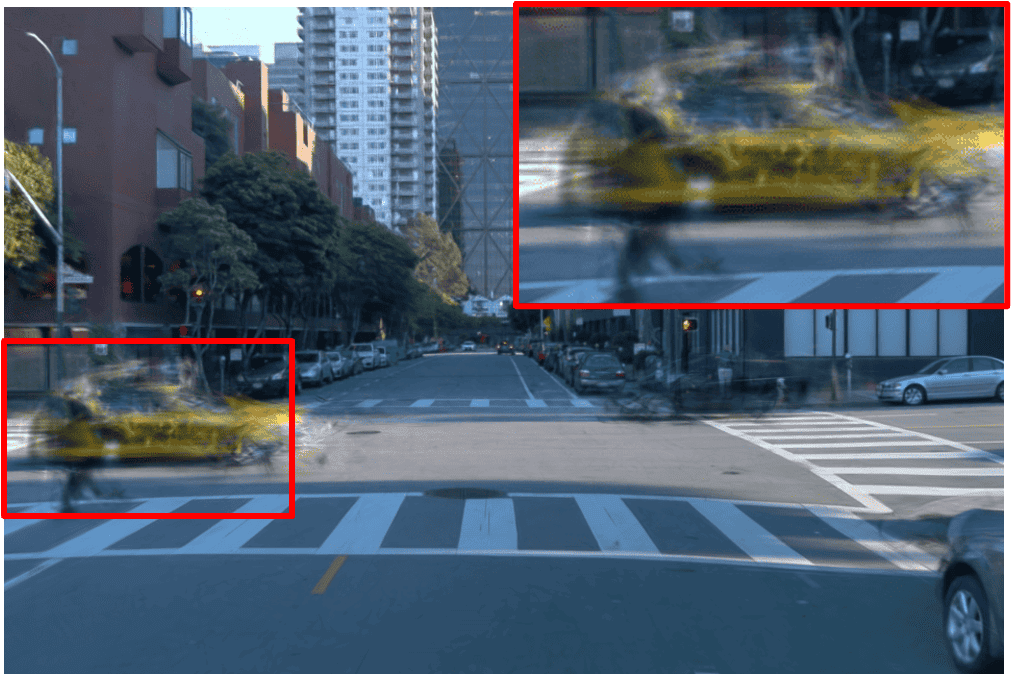}
    \caption{DeSiRe-GS~\cite{peng2024desire} (Annotation-Free)}
    \label{fig:short-desire}
  \end{subfigure}
  \begin{subfigure}{0.49\linewidth}
  \centering
    \includegraphics[width=0.49\linewidth]{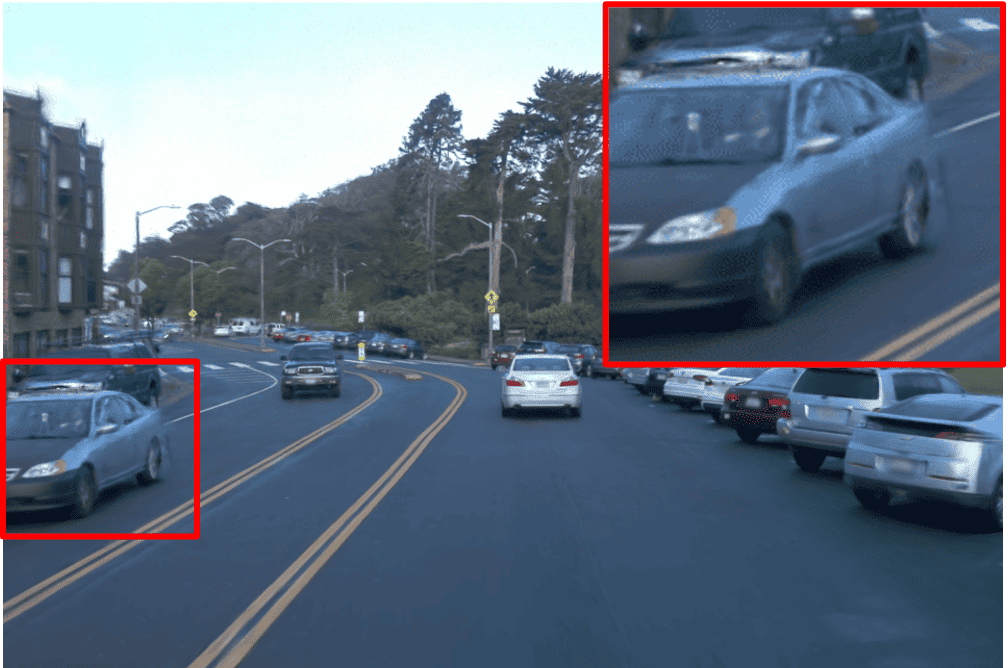}
    \includegraphics[width=0.49\linewidth]{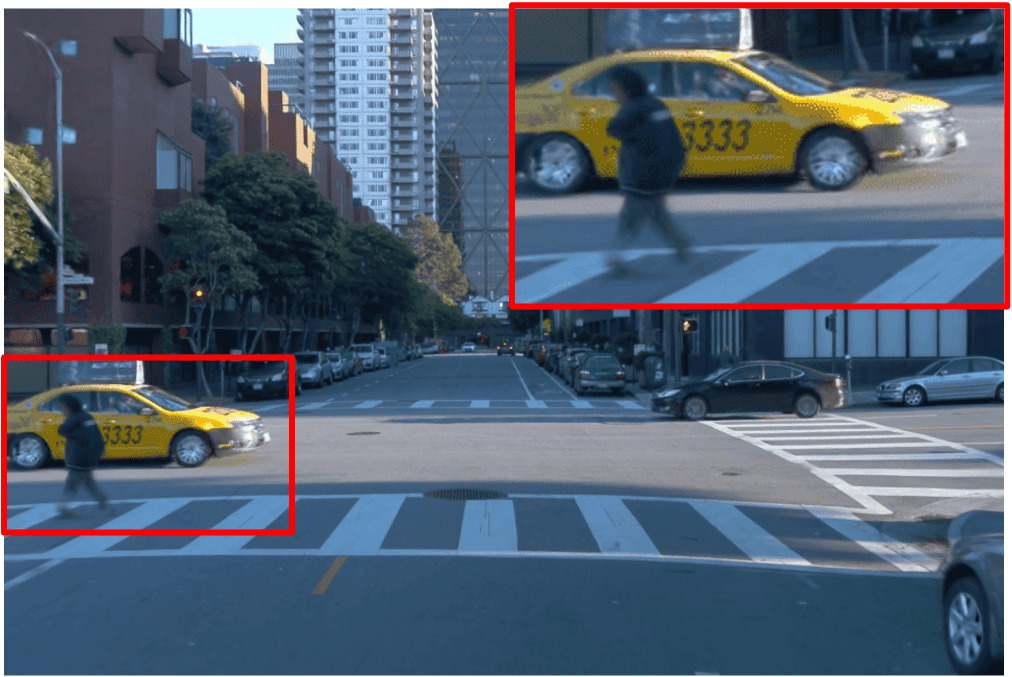}
    \caption{StreetGS~\cite{yan2024streetgs} (Annotation-Based)}
    \label{fig:short-omnire}
  \end{subfigure}
  
  \begin{subfigure}{0.49\linewidth}
  \centering
    \includegraphics[width=0.49\linewidth]{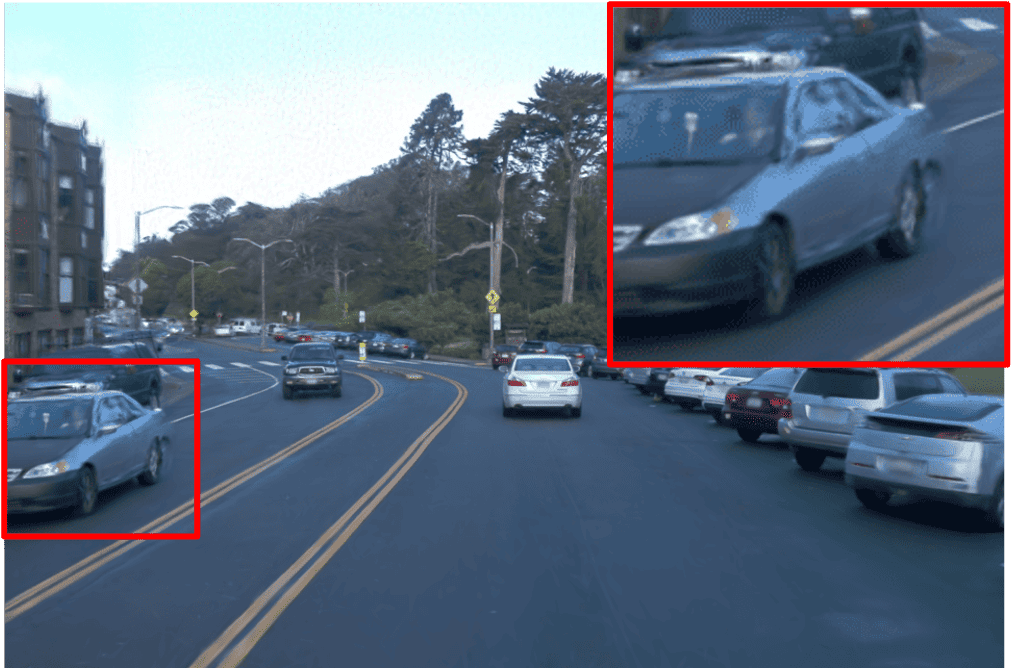}
    \includegraphics[width=0.49\linewidth]{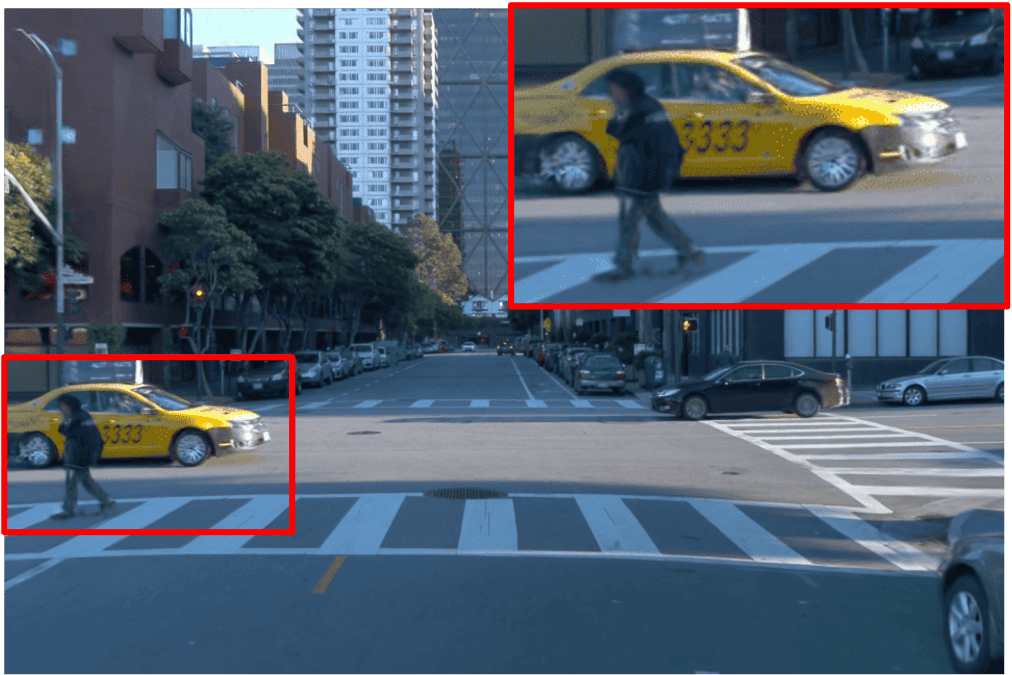}
    \caption{\textbf{Ours} (Annotation-Free)}
    \label{fig:short-Ours}
  \end{subfigure}
  \begin{subfigure}{0.49\linewidth}
  \centering
    \includegraphics[width=0.49\linewidth]{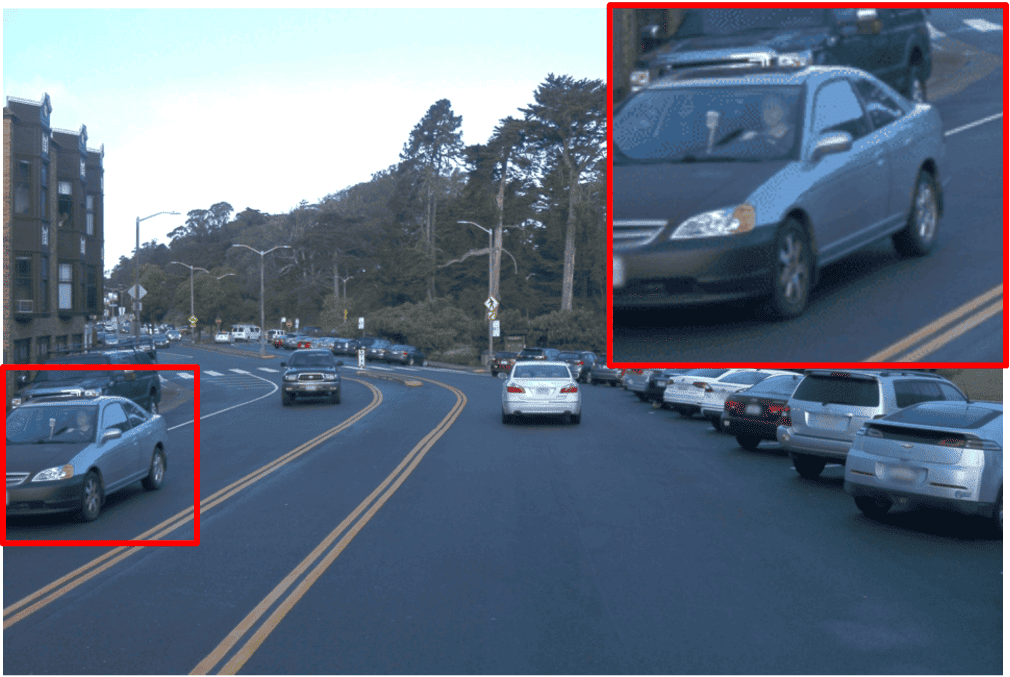}
    \includegraphics[width=0.49\linewidth]{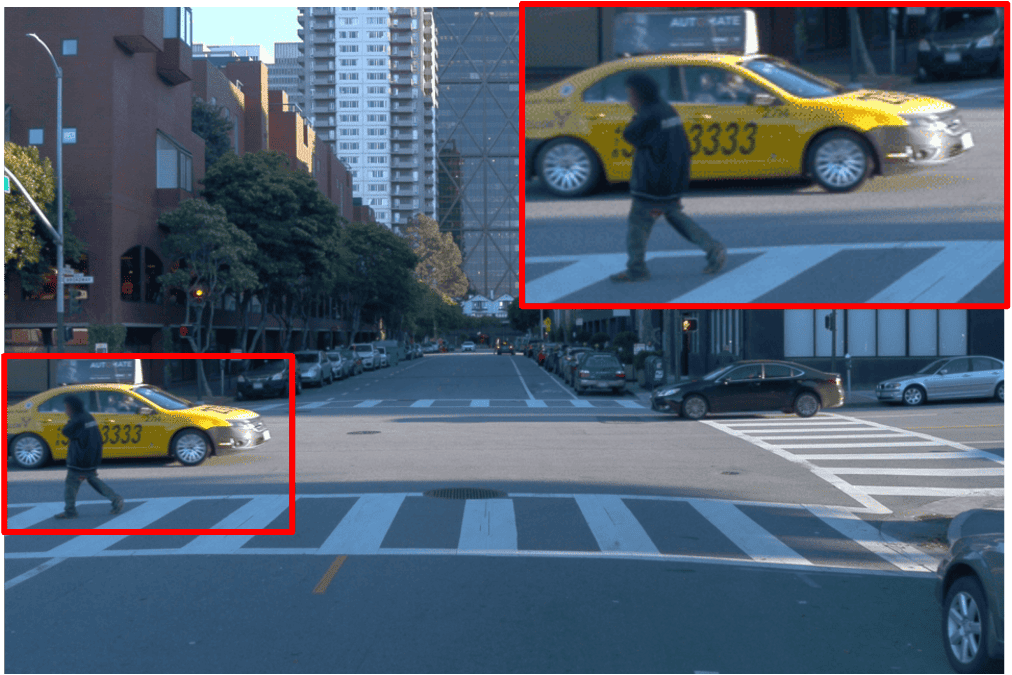}
    \caption{GT}
    \label{fig:short-GT}
  \end{subfigure}
  \vspace{-0.2cm}
  \caption{\textbf{Novel View Synthesis comparison on Waymo Open Dataset.}}
  \label{waymo_quanl}
\end{figure*}

\begin{figure}
  \centering

  \begin{subfigure}{0.49\linewidth}
  \centering
    \includegraphics[width=\linewidth]{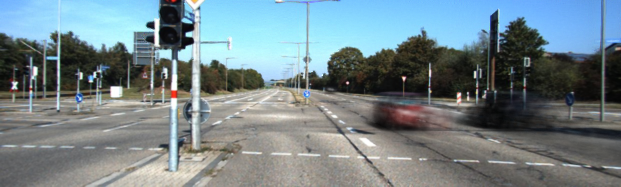}
    \caption{DeSiRe-GS~\cite{peng2024desire}}
    \label{fig:short-desire}
  \end{subfigure}
  \begin{subfigure}{0.49\linewidth}
  \centering
    \includegraphics[width=\linewidth]{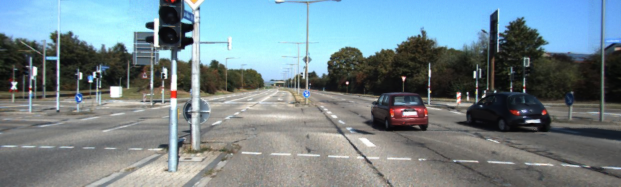}
    \caption{StreetGS~\cite{yan2024streetgs}}
    \label{fig:short-omnire}
  \end{subfigure}
  
  \begin{subfigure}{0.49\linewidth}
  \centering
    \includegraphics[width=\linewidth]{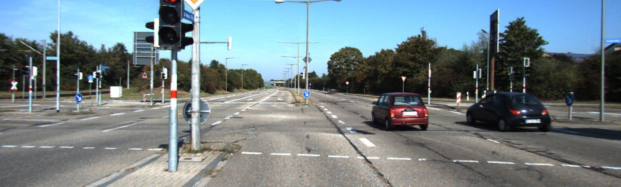}
    \caption{\textbf{Ours}}
    \label{fig:short-Ours}
  \end{subfigure}
  \begin{subfigure}{0.49\linewidth}
  \centering
    \includegraphics[width=\linewidth]{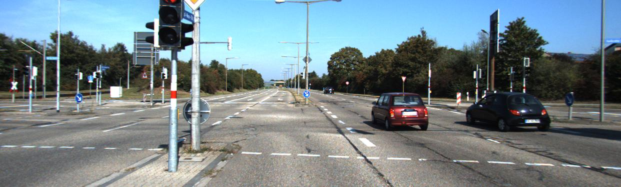}
    \caption{GT}
    \label{fig:short-GT}
  \end{subfigure}
  \vspace{-0.5cm}
  \caption{\textbf{Novel View Synthesis on KITTI Dataset.}}
  \label{kitti_quanl}
\end{figure}

\section{Experiments}

\boldparagraph{Datasets} We conduct our experiments on two real-world datasets: Waymo Open Dataset~\cite{sun2020waymo} and KITTI Dataset~\cite{Geiger2012kitti}. We use the same scene selections as OmniRe~\cite{chen2024omnire} and PVG~\cite{chen2023periodic}. Following OmniRe~\cite{chen2024omnire}, we evaluate our method on image reconstruction and novel view synthesis (NVS) tasks, using every 10th frame as the held-out test set for NVS.

\boldparagraph{Baselines} We compare our method with several state-of-the-art methods in dynamic urban scene reconstruction: EmerNeRF~\cite{yang2023emernerf}, PVG~\cite{chen2023periodic}, DeSiRe-GS~\cite{peng2024desire}, HUGS~\cite{zhou2024hugs}, StreetGS~\cite{yan2024streetgs}, and OmniRe~\cite{chen2024omnire}. Among these methods, EmerNeRF is a NeRF-based self-supervised method. PVG and DeSiRe-GS are 3DGS-based self-supervised methods that incorporate temporal variations in 3D Gaussian representations. HUGS, StreetGS, and OmniRe are scene graph-based approaches that rely on 3D bounding box annotations.

\boldparagraph{Metrics} We adopt PSNR, SSIM~\cite{wang2004image} and LPIPS~\cite{zhang2018unreasonable} as default settings for quantitative assessment of image reconstruction and novel view synthesis. Additionally, we also use PSNR and SSIM for dynamic regions, following OmniRe, to evaluate the quality of dynamic object reconstruction in the scene. For evaluating the quality of geometry reconstruction, we use Depth L1, which measures the absolute difference between the rendered depth and the ground truth obtained from projected LiDAR point clouds. 

\begin{table}[t]
\centering
\scriptsize
\begin{tabular}{lccccccc}
\toprule
 & \multicolumn{3}{c}{Image Reconstruction}&&\multicolumn{3}{c}{Novel View Synthesis} \\ 
 \cmidrule{2-4} \cmidrule{6-8}
 \makebox[0.05\textwidth]{Methods}&  \makebox[0.03\textwidth]{PSNR $\uparrow$} &  \makebox[0.03\textwidth]{SSIM  $\uparrow$} &\makebox[0.03\textwidth]{LPIPS $\downarrow$} &&  \makebox[0.03\textwidth]{PSNR $\uparrow$} &  \makebox[0.03\textwidth]{SSIM  $\uparrow$} &\makebox[0.03\textwidth]{LPIPS $\downarrow$}   \\ 
 \midrule
 \rowcolor{annobg}
\multicolumn{8}{l}{\emph{\textbf{Annotation-based methods (require GT bounding boxes)}}} \\
HUGS~\cite{zhou2024hugs} & 27.14 & 0.908 & 0.082  && 23.91  & 0.750  &  0.094 \\
StreetGS~\cite{yan2024streetgs} & 27.59 & 0.910 & 0.065  && 24.15  & 0.793  &  \cellcolor{third}0.084 \\
OmniRe~\cite{chen2024omnire} & \cellcolor{third}28.22 & \cellcolor{third}0.916 & \cellcolor{second}0.072  && \cellcolor{first}\textbf{26.52}  & \cellcolor{first}\textbf{0.881}  &  \cellcolor{first}\textbf{0.081}\\
  \midrule
   \rowcolor{annobg}
 \multicolumn{8}{l}{\emph{\textbf{Annotation-free methods}}} \\
EmerNeRF~\cite{yang2023emernerf} & 26.95 & 0.831 & 0.197  && 25.11  & 0.801  &  0.227  \\
PVG~\cite{chen2023periodic} & 27.40 & 0.895 & 0.097  && 24.34  & 0.819  &  0.121 \\
DeSiRe-GS~\cite{peng2024desire} & \cellcolor{second}28.62 & \cellcolor{second}0.921 & \cellcolor{third}0.085  && \cellcolor{third}25.32  & \cellcolor{third}0.846  &  0.096 \\

\textbf{Ours} & \cellcolor{first}\textbf{28.92} & \cellcolor{first}\textbf{0.929} & \cellcolor{first}\textbf{0.064}  && \cellcolor{second}26.10  & \cellcolor{second}0.884  & \cellcolor{second}0.079   \\
\bottomrule
% &$\checkmark$
\end{tabular}
\vspace{-0.2cm}
\caption{\textbf{Quantitative comparison on KITTI Dataset.}}
\label{kitti_quant}
\end{table}

\subsection{Experiment Results}

 % Since OmniRe relies on human-labeled ground truth bounding boxes, while other methods use predicted bounding boxes or none at all, we introduce an additional experimental settings for a fair comparison: we evaluate OmniRe with bounding boxes predicted by~\cite{casa2022, wu2021castrack}, denoted as OmniRe*. %Second, we initialize our method with ground truth bounding boxes for canonical space and per-point deformation, denoted as Ours w/ GT BBox. 

\boldparagraph{Novel View Synthesis} As shown in \cref{waymo_quant,kitti_quant}, our method not only achieves state-of-the-art performance across all rendering metrics among annotation-free methods like PVG and DeSiRe-GS, but also delivers performance comparable to OmniRe, which leverages ground-truth bounding boxes and the SMPL model at the cost of requiring extensive annotations and templates. 

Qualitative comparisons in \cref{waymo_quanl} and \cref{kitti_quanl} further demonstrate the effectiveness of our approach for reconstructing dynamic objects, whereas DeSiRe-GS often fails to preserve motion details, highlighting the benefit of canonical space. StreetGS also struggles with human reconstruction due to the lack of human-specific modeling.

In contrast, our approach achieves competitive vehicle and human reconstruction results. These results indicate that our 4D-superpoint initialization provides a strong alternative to manual annotations, and that per-point deformation offers a scalable, template-free solution for human reconstruction in urban environments.

% In the ground-truth bounding box setting, our method (Ours w/ GT BBox) outperforms OmniRe in all metrics, except for SSIM and LPIPS in novel view synthesis on full images and humans. Notably, OmniRe leverages SMPL~\cite{SMPL:2015}, a powerful human shape model, which contributes to its superior performance. These results indicate that per-point deformation is a viable solution for human reconstruction in urban scenes without requiring templates. Furthermore, our method attains a slightly higher PSNR for vehicles compared to OmniRe. This improvement is attributed to our per-point deformation mechanism, which allows for subtle deformations, such as wheel steering, leading to a more realistic reconstruction of dynamic vehicles.

\boldparagraph{Geometry} For geometry evaluation, as shown in \cref{waymo_quanl}, our method outperforms all annotation-free methods in depth L1, except for DeSiReGS, which benefits from additional supervision via normal maps predicted by a pre-trained model. This result further demonstrates that improved geometry enhances rendering quality.

\begin{table}[tb]
\centering
\scriptsize
\begin{tabular}{lcccccc}
\toprule
 & \multicolumn{2}{c}{Full PSNR $\uparrow$} & \multicolumn{2}{c}{Human PSNR $\uparrow$} & \multicolumn{2}{c}{Vehicle PSNR $\uparrow$} \\
 \cmidrule{2-7}
\makebox[0.04\textwidth]{Settings} &  \makebox[0.02\textwidth]{Recon.} &  \makebox[0.02\textwidth]{NVS} &\makebox[0.02\textwidth]{Recon.} &  \makebox[0.02\textwidth]{NVS} & \makebox[0.02\textwidth]{Recon.} &  \makebox[0.02\textwidth]{NVS} \\ 
\midrule
w/o 2D smooth  & 35.21 & 30.97 & 30.39 & 21.92 & 29.79 & 24.22 \\
w/o 3D smooth  & 35.32 & 31.34 & 30.14 & 22.61 & 30.02 & 24.32 \\
\midrule
Full model & \textbf{35.58} & \textbf{31.56} & \textbf{30.44} & \textbf{22.75} & \textbf{30.62} & \textbf{24.82} \\
\bottomrule
% &$\checkmark$
\end{tabular}
\vspace{-0.2cm}
\caption{\textbf{Ablation studies on Smooth Regularization.}}
\label{tab:ablationsmooth}
\end{table}

\begin{table}[ht]
\centering
\scriptsize
\setlength{\tabcolsep}{4.5pt}
\begin{tabular}{llccccccc}
\toprule
 && \multicolumn{3}{c}{Image Reconstruction}&&\multicolumn{3}{c}{Novel View Synthesis} \\ 
 \cmidrule{3-5}  \cmidrule{7-9}
 &\makebox[0.05\textwidth]{Init.}&  \makebox[0.03\textwidth]{PSNR $\uparrow$} &  \makebox[0.03\textwidth]{SSIM  $\uparrow$} &\makebox[0.03\textwidth]{LPIPS $\downarrow$} &&  \makebox[0.03\textwidth]{PSNR $\uparrow$} &  \makebox[0.03\textwidth]{SSIM  $\uparrow$} &\makebox[0.03\textwidth]{LPIPS $\downarrow$} \\ 
\midrule
\multirow{2}{*}{Waymo} & casa & 34.62 & 0.936 & 0.061 && 30.17 & 0.902 & 0.086 \\
& 4D-S.P. & \textbf{35.58} & \textbf{0.967} & \textbf{0.053}  && \textbf{31.56}  & \textbf{0.935}  &  \textbf{0.074} \\
\midrule
\multirow{2}{*}{KITTI} & casa & 27.15 & 0.901 & 0.087 && 25.28 & 0.861 & 0.102 \\
& 4D-S.P.  & \textbf{28.92} & \textbf{0.929} & \textbf{0.064}  && \textbf{26.10}  & \textbf{0.884}  &  \textbf{0.079} \\

\bottomrule
% &$\checkmark$
\end{tabular}
\vspace{-0.2cm}
\caption{\textbf{Ablation studies on 4D Superpoint (4D-S.P.) Initialization.} Init. indicates initialization method.}
\label{tab:ablationinit}
\end{table}

\begin{figure}
\vspace{-0.5cm}
    \centering
    \begin{subfigure}{0.32\linewidth}
        \includegraphics[width=\linewidth]{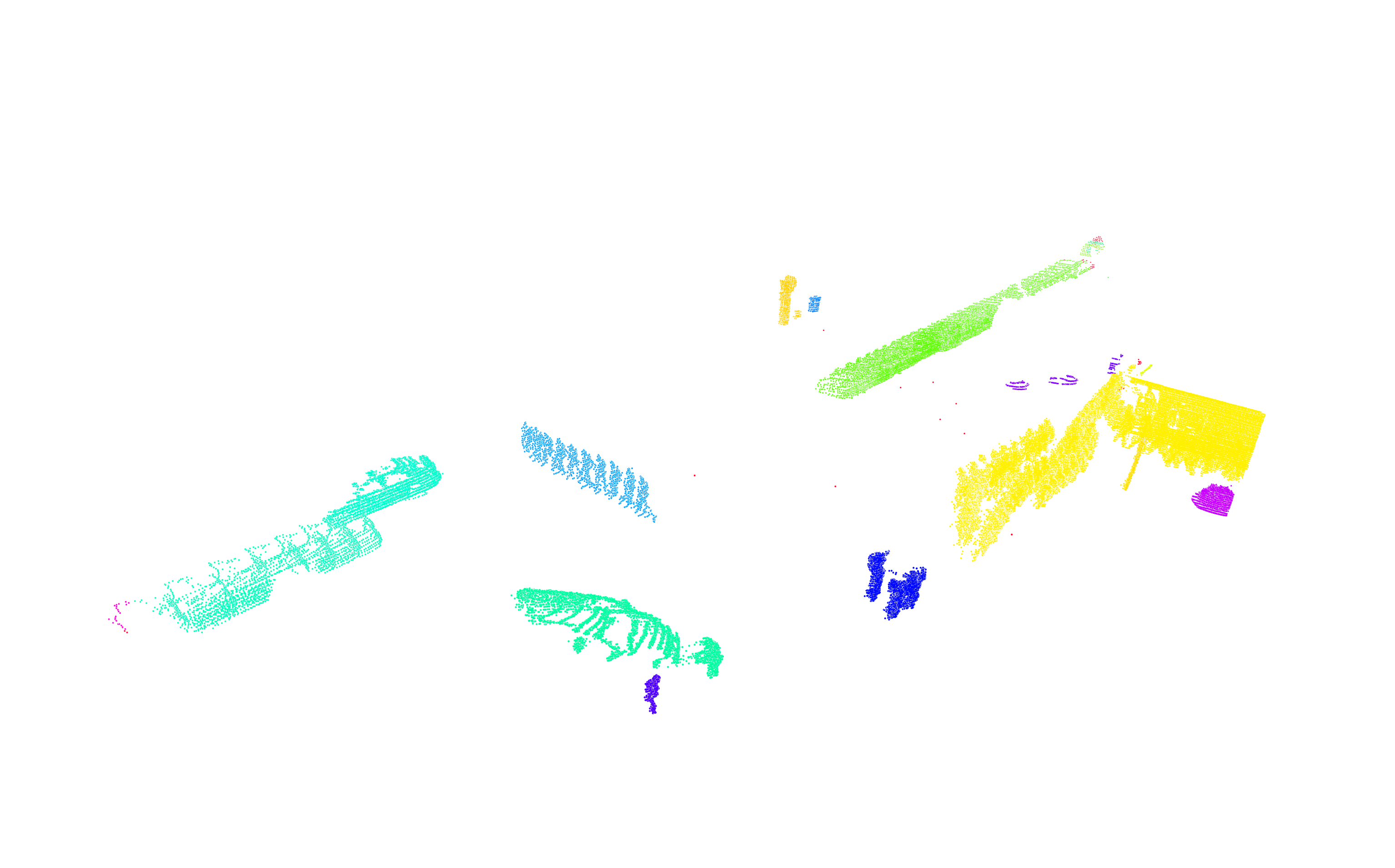}
        \caption{AS. DBSCAN}
    \end{subfigure}
    \begin{subfigure}{0.32\linewidth}
        \includegraphics[width=\linewidth]{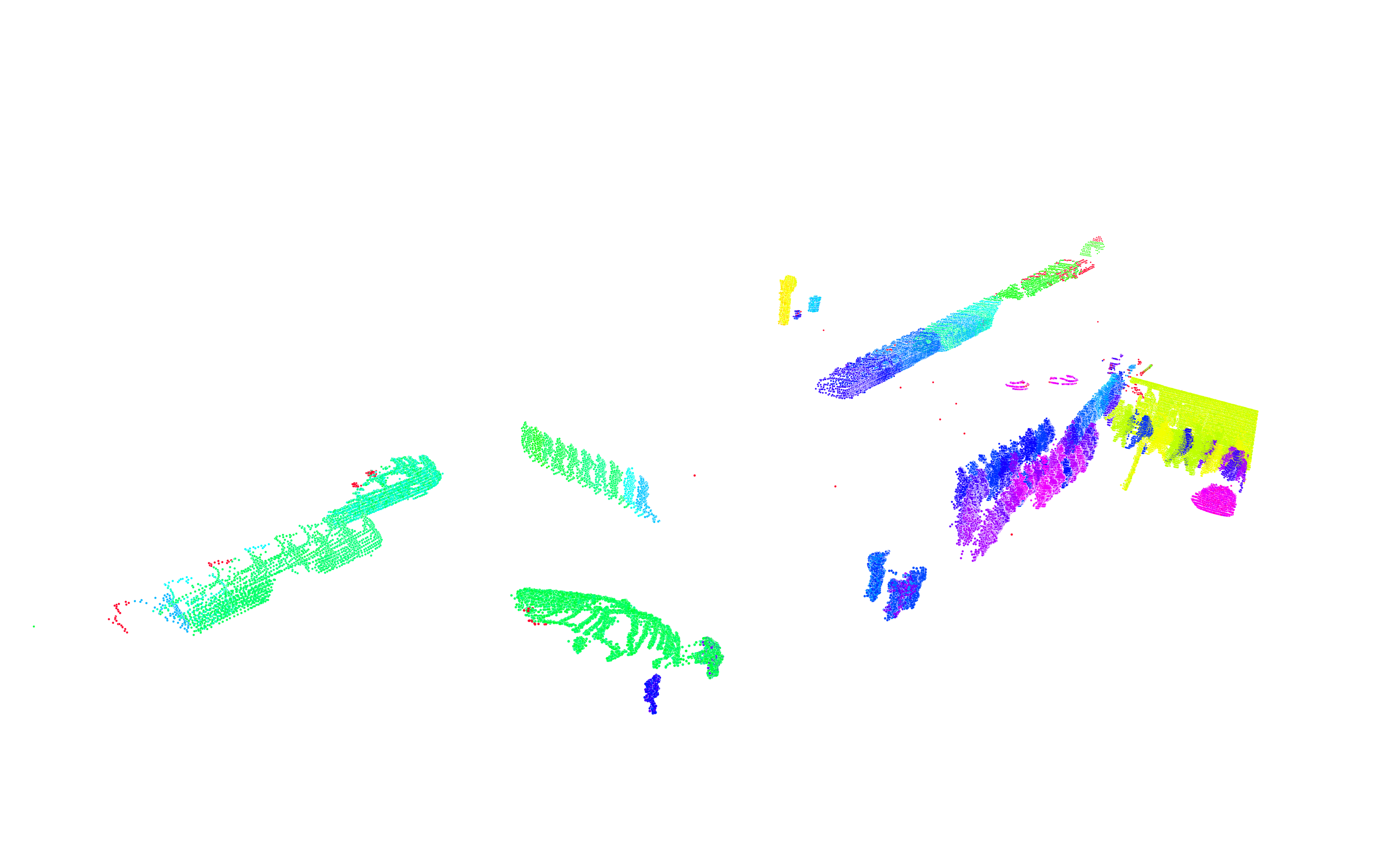}
        \caption{PF. DBSCAN}
    \end{subfigure}
    \begin{subfigure}{0.32\linewidth}
        \includegraphics[width=\linewidth]{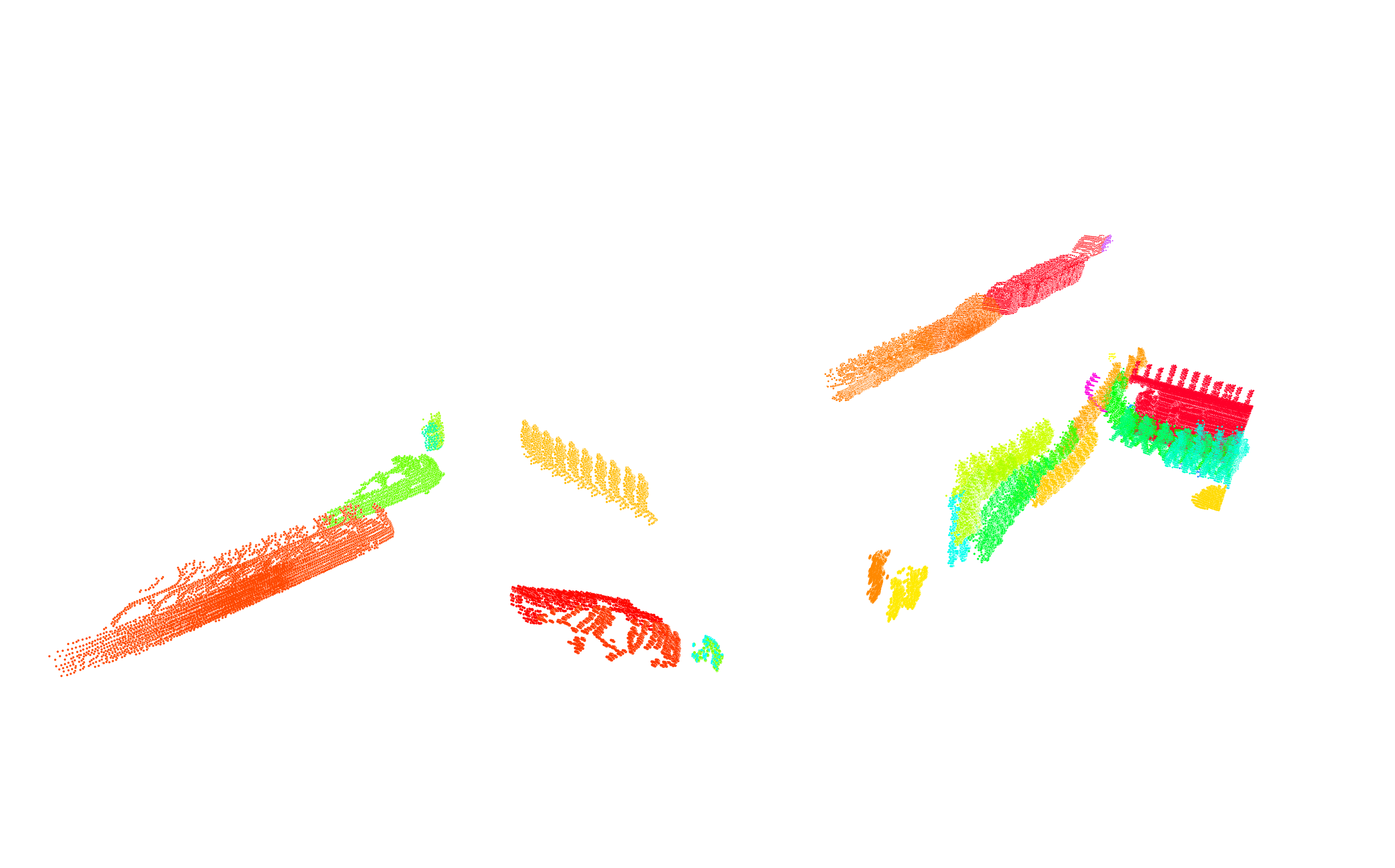}
        \caption{Ours}
    \end{subfigure}
    \vspace{-0.2cm}
    \caption{\textbf{Ablation of 4D superpoint Initialization.} AS. DBSCAN refers to applying DBSCAN clustering over the entire sequence, while PF. DBSCAN denotes performing DBSCAN independently on each frame throughout the sequence.}
    \label{fig:ablateinit}
\end{figure}

\subsection{Ablation Study}

To verify the effectiveness of our proposed methods, 
we conduct ablation studies on the Waymo and KITTI datasets. 
%The results are listed in \cref{tab:ablationsmooth}. 
% More ablation results are provided in~\cref{sec:supp_results}
% .

\boldparagraph{4D Superpoint-Based Initialization}
We first evaluate the impact of our 4D initialization by replacing it with bounding boxes predicted by casa~\cite{casa2022, wu2021castrack} for canonical space and per-point deformation. 
As shown in \cref{tab:ablationinit}, compared with initialization with predicted bounding boxes, our 4D superpoint-based initialization improves reconstruction accuracy. This highlights the role of our initialization to provide a good prior to the scene representation and preserve high-quality reconstructions across large urban environments. Next, we compare our 4D initialization against naive instance clustering methods. As shown in \cref{fig:ablateinit}, applying DBSCAN over the entire sequence leads to under-segmentation, while per-frame DBSCAN fails to maintain consistent instance IDs across frames, resulting in inconsistent instance tracking. In contrast, our method effectively decomposes individual instances throughout the sequence. 

%We first evaluate the impact of our 4D initialization by replacing it with predicted bounding boxes for canonical space and per-point deformation (w/o Init.), similar to OmniRe*. As shown in \cref{tab:ablationinit}, our 4D superpoint-based initialization improves reconstruction accuracy, demonstrating its role in providing a strong prior for scene representation and maintaining high-quality reconstructions in large urban environments. Next, we compare our 4D initialization with naive instance clustering methods. As shown in \cref{fig:ablateinit}, our approach effectively decomposes individual instances across the sequence. In contrast, applying DBSCAN over the entire sequence leads to under-segmentation, and per-frame DBSCAN fails to maintain consistent instance IDs across frames, resulting in inconsistent tracking.

\begin{figure}
    \centering
    \vspace{-0.2cm}
    \begin{subfigure}{0.32\linewidth}
        \includegraphics[width=\linewidth]{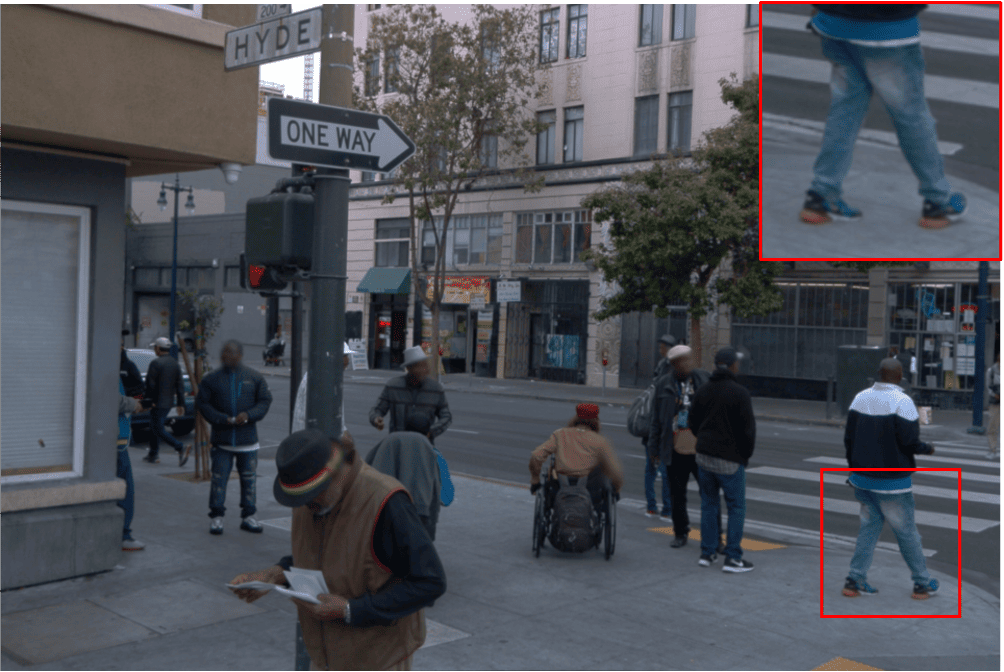}
        \caption{GT}
    \end{subfigure}
    \begin{subfigure}{0.32\linewidth}
        \includegraphics[width=\linewidth]{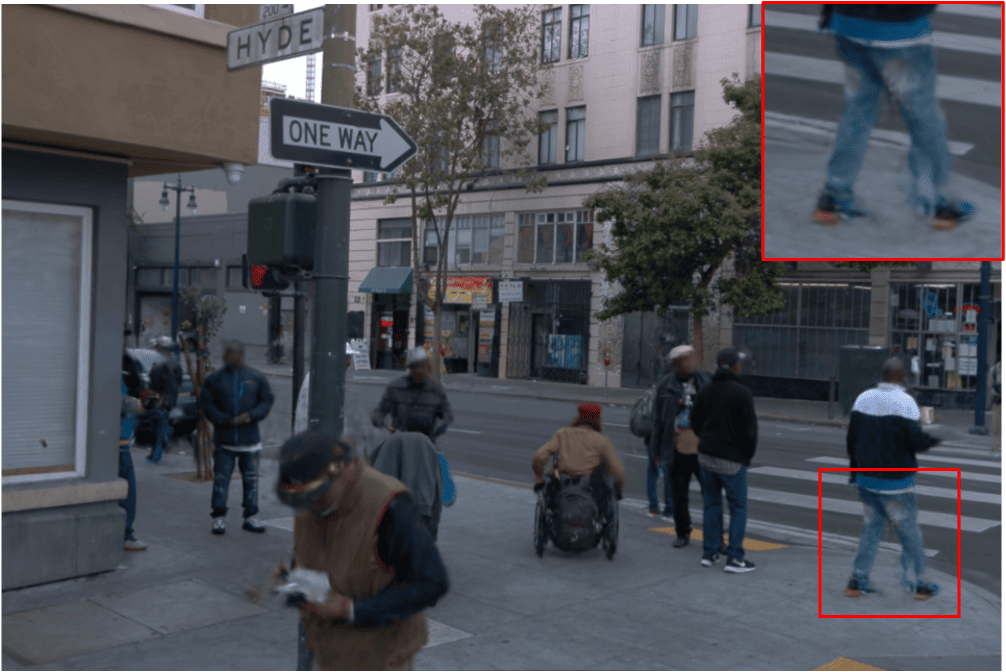}
        \caption{w/o 2D s.r.}
    \end{subfigure}
    \begin{subfigure}{0.32\linewidth}
        \includegraphics[width=\linewidth]{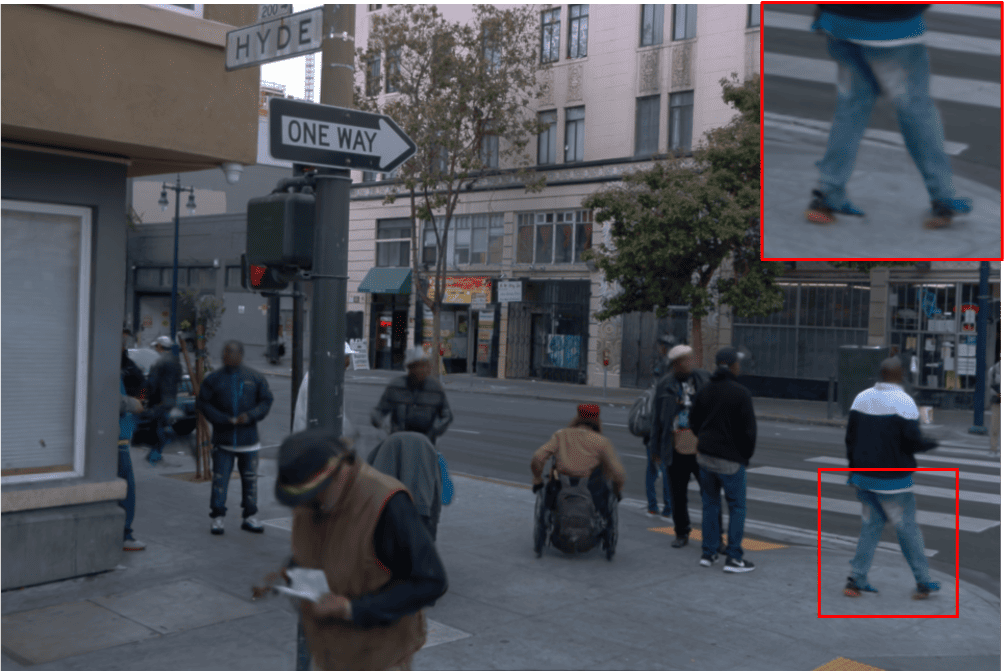}
        \caption{w/ 2D s.r.}
    \end{subfigure}
    \vspace{-0.2cm}
    \caption{\textbf{Ablation of 2D smoothness regularization (s.r.).}}
    \label{ablation_smooth}
\end{figure}

\boldparagraph{Temporal Smoothness Regularization} To evaluate the effect of smoothness regularization, we conduct experiments with and without the 2D and 3D smoothness losses. Results in \cref{ablation_smooth} and \cref{tab:ablationsmooth} show that removing smoothness losses leads to unstable motion trajectories in novel view synthesis (NVS) and reduced motion consistency across frames. This ablation experiment emphasizes the importance of smoothness regularization in ensuring temporal consistency and improving generalization in dynamic scene reconstruction.

\begin{figure}
    \centering
    \begin{subfigure}{0.32\linewidth}
        \includegraphics[width=\linewidth]{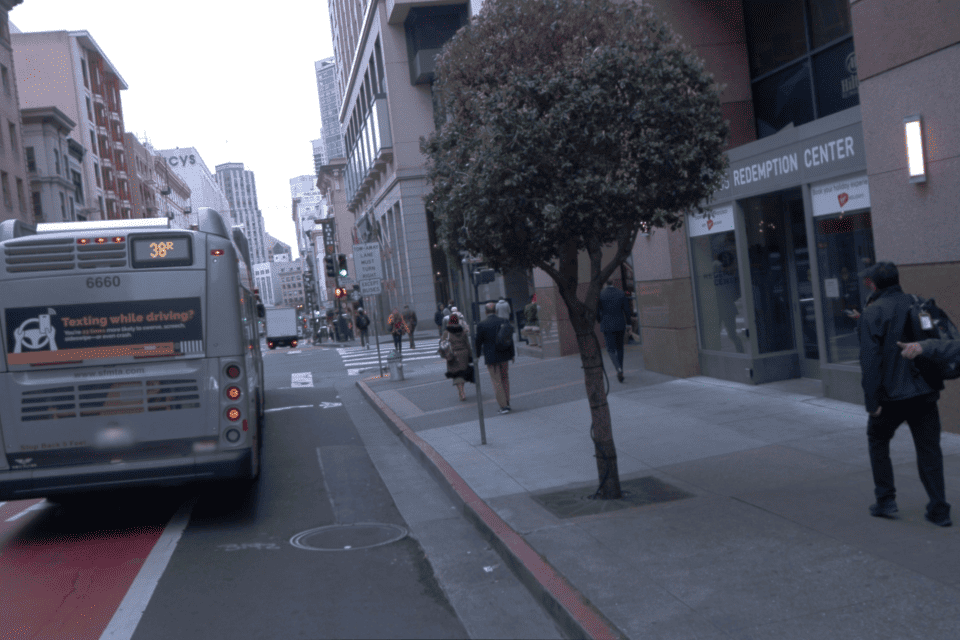}
        \includegraphics[width=\linewidth]{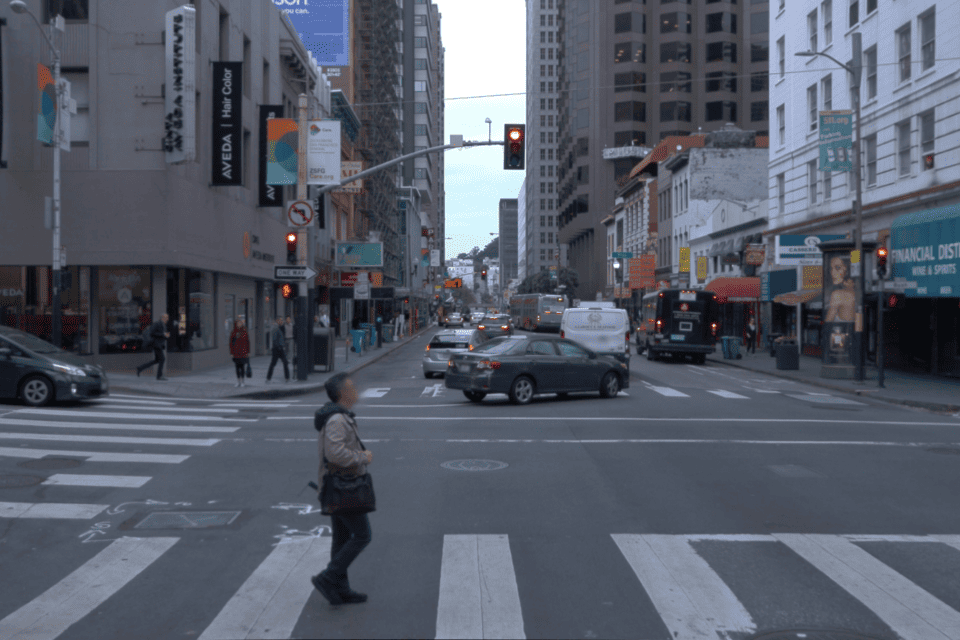}
        \caption{GT}
    \end{subfigure}
    \begin{subfigure}{0.32\linewidth}
        \includegraphics[width=\linewidth]{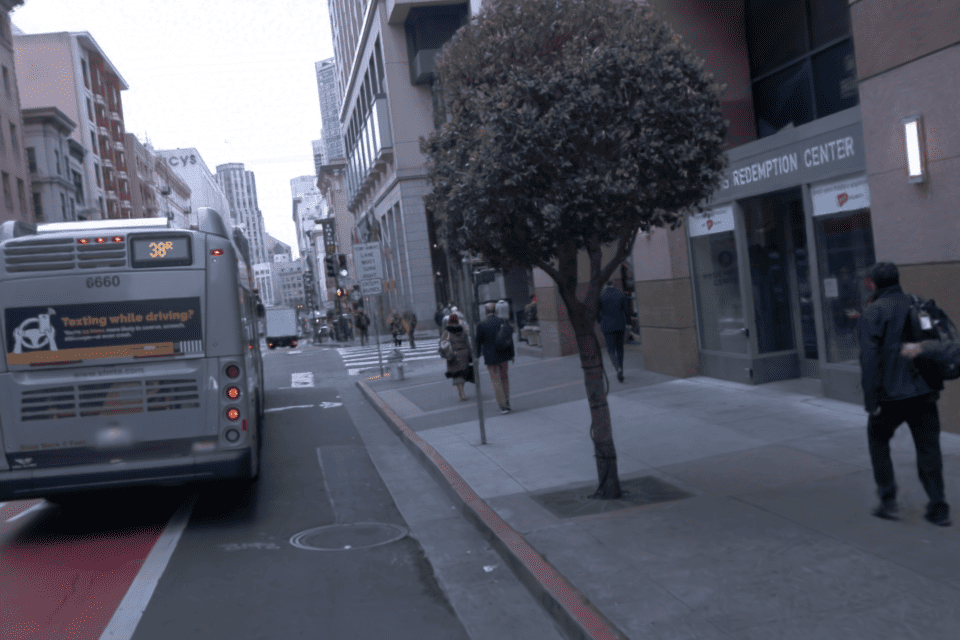}
        \includegraphics[width=\linewidth]{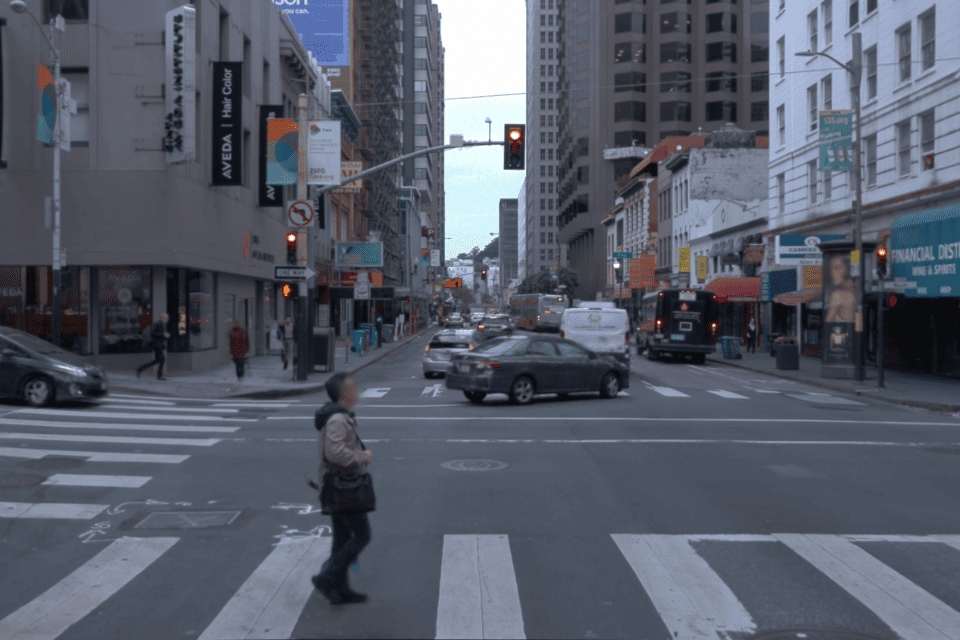}
        \caption{Rendering}
    \end{subfigure}
    \begin{subfigure}{0.32\linewidth}
        \includegraphics[width=\linewidth]{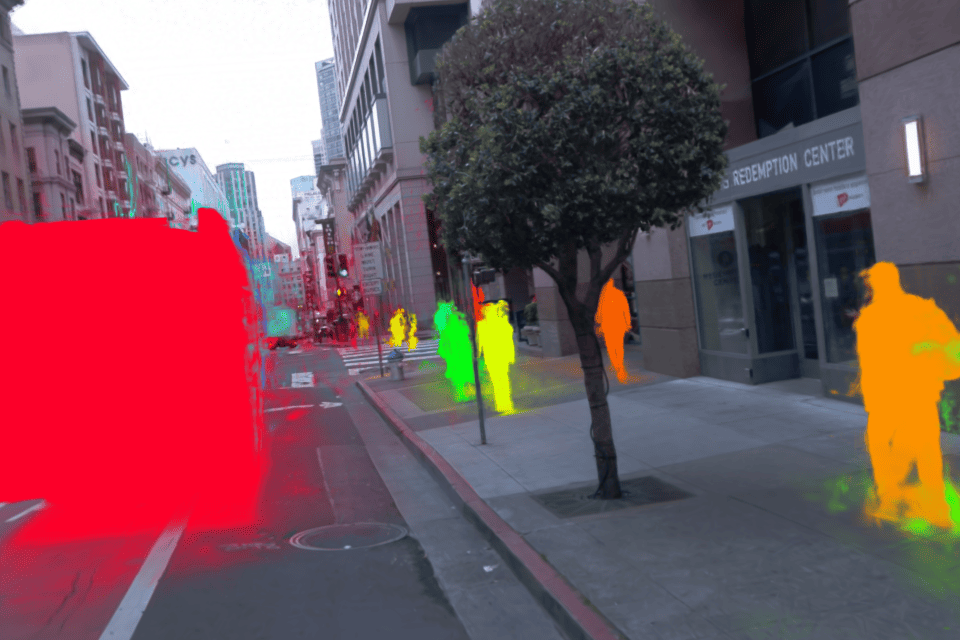}
        \includegraphics[width=\linewidth]{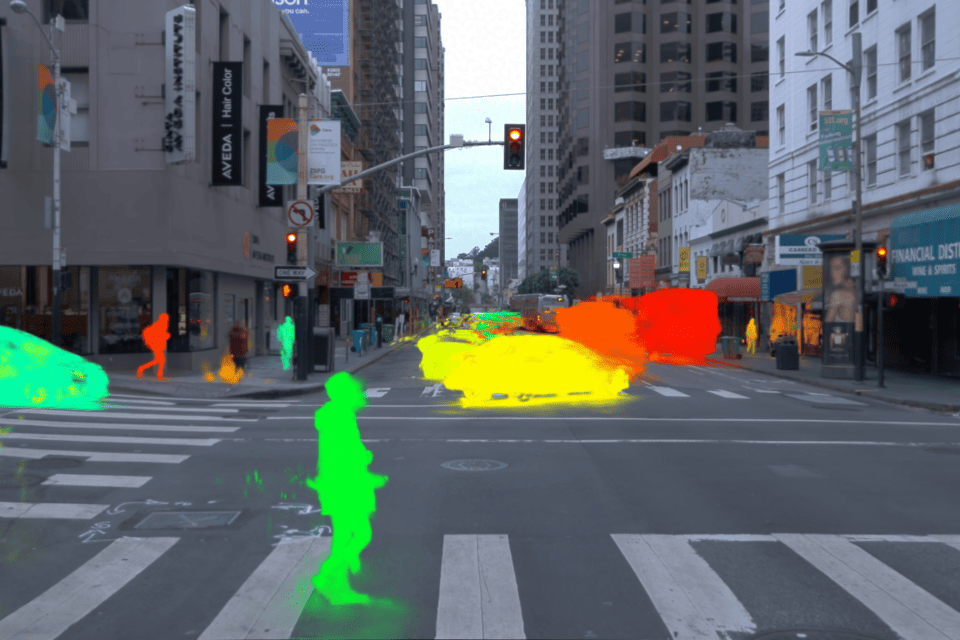}
        \caption{Decomposition}
    \end{subfigure}
    \vspace{-0.2cm}
    \caption{\textbf{Visualization of dynamic instance decomposition.}}
    \label{fig:dynamic_instance_seg}
\end{figure}

\subsection{Application}

\boldparagraph{Dynamic Instance Decomposition}
Our method excels in the decomposition of dynamic objects in urban environments, offering robust and temporally consistent instance identification across a sequence of frames. By leveraging unsupervised 4D initialization, we achieve accurate and scalable instance decomposition in dynamic urban scene. As shown in \cref{fig:dynamic_instance_seg}, UnIRe successfully decomposes vehicles, pedestrians, and other dynamic objects.

\begin{figure}
    \centering
    % \begin{subfigure}{0.24\linewidth}
    %     \includegraphics[width=\linewidth]{img/application/edit/origin.png}
    %     \caption{Origin}
    % \end{subfigure}
    % \begin{subfigure}{0.75\linewidth}
    %     \includegraphics[width=0.32\linewidth]{img/application/edit/t1.png}
    %     \includegraphics[width=0.32\linewidth]{img/application/edit/t2.png}
    %     \includegraphics[width=0.32\linewidth]{img/application/edit/t3.png}
    %     \caption{Editing}
    % \end{subfigure}
    \includegraphics[width=\linewidth]{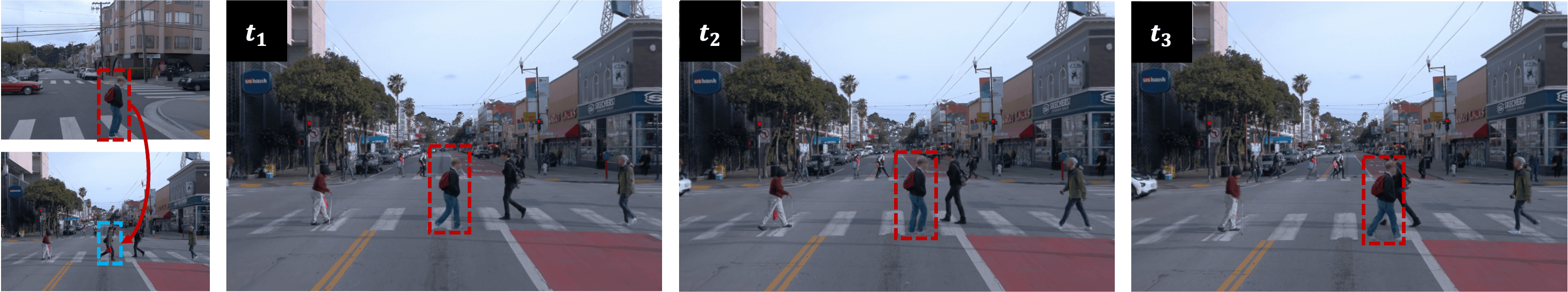}
    % \vspace{-0.6cm}
    \caption{\textbf{Scene Editing.} An example of scene editing, where a pedestrian is replaced with another in a different scene.}
    \label{fig:scene_editing}
\end{figure}

\boldparagraph{Scene Editing}
Beyond instance decomposition, our method is also capable of manipulating dynamic urban scenes through scene editing. This includes operations such as object removal, replacement, and motion editing, all while preserving the overall scene structure and consistency. As illustrated in \cref{fig:scene_editing}, our approach allows for editing of dynamic scenes, enabling realistic modifications with minimal artifacts. This demonstrates the potential of UnIRe for applications in AR/VR, and autonomous driving simulations.

\section{Conclusion}

In this paper, we introduce UnIRe, a 3DGS-based approach that decomposes a scene into a static background and individual dynamic instances using only RGB images and LiDAR point clouds, eliminating the need for bounding boxes or object templates.
By incorporating 4D superpoints, a novel representation that clusters multi-frame LiDAR points in 4D space, UnIRe facilitates unsupervised instance separation through spatiotemporal correlations, leading to state-of-the-art performance in image reconstruction and enabling instance-level editing.
% {
%     \small
%     \bibliographystyle{ieeenat_fullname}
%     \bibliography{main}
% }
\bibliographystyle{IEEEtran}
\bibliography{IEEEabrv,refs}

% WARNING: do not forget to delete the supplementary pages from your submission 
% \input{sec/X_suppl}

\end{document}